\documentclass[preprint,authoryear,3p]{elsarticle}

\usepackage{booktabs}

\usepackage{amssymb,amsthm,amsmath}
\usepackage{graphics}
\usepackage{graphicx}
\usepackage{subfig}
\usepackage{float}
\usepackage{multirow}
\usepackage{longtable,lscape}
\usepackage{cases}
\usepackage{url}
\usepackage{caption}
\usepackage{color}
\usepackage{amsthm}
\usepackage[pagewise]{lineno}
\usepackage{mathtools}
\usepackage{url}
\usepackage[colorlinks]{hyperref}

\usepackage{fancyhdr}
\usepackage{setspace}

\usepackage{nomencl}
\makenomenclature

\usepackage{etoolbox}
\renewcommand\nomgroup[1]{%
 \item[\bfseries
  \ifstrequal{#1}{A}{Reinforcement Learning}{%
  \ifstrequal{#1}{B}{Traffic Signal Control}{%
  \ifstrequal{#1}{C}{Other Notations}{}}}%
]}

\usepackage{array}
\usepackage{multirow}
\usepackage{multicol}
\usepackage{makecell}
\usepackage{comment}
\usepackage{graphicx}
\usepackage{soul}

\usepackage{booktabs,caption}
\usepackage{threeparttable} %[flushleft]
\usepackage{arydshln}
\usepackage{xcolor}
\usepackage[ruled,vlined, linesnumbered]{algorithm2e}

\usepackage{enumerate}

\usepackage{makecell}%To keep spacing of text in tables
\setcellgapes{4pt}%parameter for the spacing
\usepackage[export]{adjustbox}
\newcommand{\tcb}[1]{{\textcolor{black}{#1}}}

\usepackage{fullpage}

\theoremstyle {plain}% default

\newcounter{x}

\theoremstyle{definition}

\theoremstyle{remark}

\usepackage{soul}
\usepackage{titlesec}
\titleformat{\paragraph}
{\normalfont\normalsize\bfseries}{\theparagraph}{1em}{}
\titlespacing*{\paragraph}
{0pt}{3.25ex plus 1ex minus .2ex}{1.5ex plus .2ex}

\usepackage{color}

\makeatletter
\def\ps@pprintTitle{%
	\let\@oddhead\@empty
	\let\@evenhead\@empty
	\def\@oddfoot{\reset@font\hfil\thepage\hfil}
	\let\@evenfoot\@oddfoot
}
\makeatother

\numberwithin{equation}{section}

\journal{Transportation Research Part C Special Issue ``Managing Future Motorway and Urban Traffic Systems"}

\begin{document}

\begin{frontmatter} 
		
		

\noindent 
\textcolor{blue}{Published in: \href{https://doi.org/10.1016/j.trc.2022.103728} {{\color{blue} \textit{Transportation Research Part C: Emerging Technologies} 141 (2022): 103728}}.
Please cite this paper as: \\Zhaobin Mo, Wangzhi Li, Yongjie Fu, Kangrui Ruan, and Xuan Di. "CVLight: Decentralized learning for adaptive traffic signal control with connected vehicles." \textit{Transportation Research Part C: Emerging Technologies} 141 (2022): 103728. DOI:\\ \href{https://doi.org/10.1016/j.trc.2022.103728}{{\color{blue} https://doi.org/10.1016/j.trc.2022.103728}}}

\title{
\textbf{CVLight}: Decentralized Learning for Adaptive Traffic Signal Control with Connected Vehicles
}
		
\date{\today}
		

 \author[cu]{Zhaobin Mo\corref{cor1}}
 \author[dsi]{Wangzhi Li\corref{cor1}}
 \author[cu]{Yongjie Fu}
 \author[cu]{Kangrui Ruan}
 \author[cu,dsi]{Xuan Di\corref{cor}}
% \ead{sharon.di@columbia.edu}

\cortext[cor1]{Zhaobin Mo and Wangzhi Li contributed equally to this work.}
\cortext[cor]{Corresponding author. Tel.: +1 212 853 0435.}

\address[cu]{Department of Civil Engineering and Engineering Mechanics, Columbia University}
\address[dsi]{Data Science Institute, Columbia University}
%\address[sjtu]{Department of Computer Science and Engineering, Shanghai Jiao Tong University}
%\address[cu-ee]{Department of Electrical Engineering, Columbia University}

\begin{abstract}
{This paper develops a decentralized reinforcement learning (RL) scheme for multi-intersection adaptive traffic signal control (TSC), called ``CVLight", that leverages data collected from connected vehicles (CVs). The state and reward design facilitates coordination among agents and considers travel delays collected by CVs. A novel algorithm, Asymmetric Advantage Actor-critic (Asym-A2C), is proposed where both CV and non-CV information is used to train the critic network, while only CV information is used to execute optimal signal timing. Comprehensive experiments show the superiority of CVLight over state-of-the-art algorithms under a 2-by-2 synthetic road network with various traffic demand patterns and penetration rates. The learned policy is then visualized to further demonstrate the advantage of Asym-A2C. A pre-train technique is applied to improve the scalability of CVLight, which significantly shortens the training time and shows the advantage in performance under a 5-by-5 road network. A case study is performed on a 2-by-2 road network located in State College, Pennsylvania, USA, to further demonstrate the effectiveness of the proposed algorithm under real-world scenarios. Compared to other baseline models, the trained CVLight agent can efficiently control multiple intersections solely based on CV data and achieve the best performance, especially under low CV penetration rates.} %\tco{just "is low."}

\begin{keyword}
	Traffic signal control, 
    Deep reinforcement learning, 
    Connected vehicles,
    Actor-critic algorithm
    %Centralized Training and Decentralized Execution
\end{keyword}

\end{abstract}
		
\end{frontmatter}

\section{Introduction}
\label{sec:introduction}

\emph{Connected vehicle} (CV) refers to the vehicular technology that enables vehicle-to-vehicle (V2V) (or vehicular ad-hoc networks (VANETs) and vehicle-to-infrastructure (V2I) communication \citep{usdot_cv,di2021survey}. The enabling technology includes dedicated short-range communication (DSRC)~\citep{dsrcAndCell}, or cellular communication like 5G~\citep{IoT5GSur16,IoT5GSur20,IoT5Gsur18}. Connected vehicles will likely generate terabytes of streaming data daily~\citep{cv_sas}, holding great potential for various transportation applications including traffic signal control (TSC), which is the focus of this paper.

% \subsection{Problem Statement}
% \label{subsec:problem_statement}

% This subsection will:

% - describe the main problem this paper aims to solve. \tcr{wz: define 'low penetration rate' and 'partial CV data' here.}

% - describe the significance and specialty of this problem (how does our scope be different from others and why is this angle important?)

\subsection{Literature Review}
\label{subsec:literatrue_review}

CV can provide more detailed traffic information, real-time vehicle trajectories for example, when compared to traditional detectors. 
To incorporate CV data into TSC systems, researchers have developed various TSC strategies in recent years ~\citep{liu2014cooperative,liu2017distributed,li2016traffic,li2018connected,kim2019real,hussain2020optimizing,li2020connected,yan2020efficiency}. 
%{insert Stephen Smith's work here}
Although these proposed TSC strategies with CV data significantly outperform traditional traffic signal control methods such as actuated TSC or fixed-time TSC, usually a 100\% CV market penetration rate is assumed. 
Such a full market may not be achievable in the short term, thus motivating researchers to develop TSC strategies capable of running on limited CV data, specifically for low CV market penetration (below 30\%) scenarios. 
In this section, we will mainly focus on TSC strategies that assume only a portion of vehicles are CVs and categorize them into \textbf{non-learning based} and \textbf{reinforcement learning (RL) based} TSC strategies. Readers who are interested in a more comprehensive overview of traffic control with connected vehicles can refer to \cite{guo2019urban}.

%Although a sufficient amount of historical traffic data is critical to RL-TSC systems, exploding volume of trajectory data are generated from various sources today, e.g. connected and automated vehicles (CAVs), navigation applications, and ride-hiring platforms \citep{wei2019survey}.

\subsubsection{Non-learning based TSC with Partial CV Information}
\label{subsubsec:review_opimization-based_TSC}

%{wangzhi: in this section, we need to stress two weaknesses of proposed optimization-based TSC systems: 1) the optimization problem can be complicated and hard to solve in real-time; 2) estimation methods are required under low penetration rate scenarios, which require high reliability of the estimation methods and acceptable assumptions.}

%The generic traffic signal control problem can be formulated as an optimization.
One major non-learning based TSC strategy includes formulating the TSC problem as an optimization problem \citep{li2013survey}.
At each time step, a TSC system receives environmental inputs (such as new vehicle arrivals)
%generating newly arrival vehicles
and updates traffic states (such as queue lengths) that can be partially or fully observable. 
%\tcr{(what is environment input? How to generate?)}
The objective of the TSC system is to optimize some performance measurements (such as total vehicle delays) over a finite period of time. 
The output is a series of control signals to the TSC system that are usually represented by traffic signal phase sequences and durations.
%\tcr{(Introducing the formulation of the optimization problem and the solution to it.)} 
The most common optimization methods for TSC problems are mixed-integer linear programming (MILP) \citep{he2012pamscod} and dynamic programming (DP) \citep{feng2015real,feng2016connected,feng2018real,hu2019cooperative}. \cite{beak2017adaptive} further develop a 2-level optimization framework: at the intersection-level a DP is used to optimize the individual vehicle delays, and at the corridor level an MILP is solved to minimize platoon delays. Greedy rules like a longest-queue-first algorithm are also adopted to solve the TSC problem. \citep{lee2013cumulative,goodall2013traffic}.
%Unlike optimization-based TSC, several non-learning-based TSC methods are similar to the longest-queue-first algorithm: greedily choosing the signal timing plan according to a certain measurement of performance, such as cumulative travel time \citep{lee2013cumulative} and cumulative vehicle delay \citep{goodall2013traffic}.

%\tcr{(I am lost in the below paragraph.)}{(the goal of this paragraph is to introduce the importance of estimation methods)}The performance of an non-learning-based TSC system is highly relevant with the detail level of the state variables and the accuracy of objective measurements. Therefore, it is promising to incorporating detailed CV data into TSC system; however, such benefits of CV data require a relatively high CV market penetration rate. 
To provide accurate state variables and objective measurements under low CV penetration rates, various traffic state estimation methods have been developed. 
They are categorized into \emph{vehicle-level} and \emph{flow-level} estimation methods based on information granularity. 
Vehicle-level algorithms estimate trajectories of each vehicle, while flow-level ones focus on aggregate traffic measurements such as traffic volumes. 
%\tco{(Delete from "Some estimation" to the end of this paragraph.)}Some estimation methods rely on more specific information such as the CV penetration rate~\citep{he2012pamscod,tiaprasert2015queue}, require certain traffic arrival patterns like Poisson distribution~\citep{zheng2017estimating}, or depend on a prior assumption that no queues exist if no CVs are observed \citep{tiaprasert2015queue}. 

For vehicle-level estimation, a majority of studies use CVs to infer positions and speeds of non-CVs~\citep{goodall2014microscopic,feng2015real,feng2016connected,beak2017adaptive}. \cite{goodall2013traffic} propose a predictive microscopic simulation algorithm (PMSA), which optimizes traffic signal timing based on simulated CVs over a time horizon from a microscopic numerical simulator. 
Without inference of non-CVs, the proposed TSC system (PMSA) requires at least a 50\% penetration rate to outperform a conventional TSC system. %\tco{(did you mention actuated traffic control?)}. 
\cite{goodall2014microscopic} further estimate unconnected vehicles (denoted by non-CVs) positions to improve the performance of PMSA under low penetration rates: stopping non-CVs are inserted into detected gaps among CVs and added to simulation with CVs.  \cite{feng2015real,feng2016connected} first divide road segments near intersections into three regions based on vehicle status, namely, a queuing region, a slow-down region, and a free-flow region. Then locations and speeds of individual non-CVs are estimated within each region. %for the CV and non-CV arrival flow prediction. 

% we can try to summarize the pros and cons in using different detail level of infomration.

Flow-level estimation infers queue lengths \citep{priemer2009decentralized,tiaprasert2015queue, li2020backpressure},  platoon arrival time \citep{he2012pamscod}, vehicle densities \citep{mohebifard2018real,mohebifard2019cooperative,al2020real}, cumulative travel time \citep{lee2013cumulative}, or traffic volumes \citep{zheng2017estimating,feng2018real}. 
%Accordingly,  linear regression \citep{he2012pamscod}, Kalman filter \citep{lee2013cumulative}, and least-mean-square-error method \citep{tiaprasert2015queue} have been applied to studies along this line. 
{Specifically, queue estimation for signalized intersections has gained more attention from researchers in recent years \citep{tiaprasert2015queue,hao2015long,yang2018queue, gao2019connected,li2020backpressure}, especially by utilizing vehicle trajectory data \citep{hao2015long,yang2018queue}.} 
Most recently, some studies focus on extremely low (below 10\%) penetration rate scenarios. \cite{feng2018real} apply a traffic volume estimation method \citep{zheng2017estimating} to an adaptive TSC system developed in \cite{feng2015real}. This traffic volume estimation method utilizes trajectory data from CVs or navigation devices to estimate current traffic volumes under a penetration rate as low as 10\%. To cope with low penetration rate scenarios, \cite{al2020real} propose 
%\tcr{(still see switch between the past and the present tense.)}
two kinds of traffic state estimation algorithms: one uses combined data from CVs and loop detectors to estimate the non-CV trajectories based on a car-following model, the other converts vehicle detection time to spatial distributions of CVs and non-CVs. Results of experiments on a real-world corridor of 4 intersections with real-world traffic demands show that both estimation methods function well under 10\% or lower penetration rates. 
%Building on both estimation methods, the proposed optimization based signal control reduces travel time and delay under 10\% or lower penetration rates in a real-world corridor of 4 intersections with real-world demand. %, compared to the existing signal plan.

Table \ref{tab:literature_review_optimization_tsc} summarizes recent studies on non-learning based TSC with partial observation (i.e. without a 100\% CV penetration rate). Each row represents the reference reviewed above, containing details of their information source, environment, estimation granularity, estimation methods, benchmarks, required minimum penetration rate, key assumptions, and gap, respectively.

In summary, when it comes to using CV data to estimate traffic states, non-learning based TSC methods usually rely on traffic models or certain assumptions of traffic. Because the performance of such TSC systems highly depends on the accuracy of traffic state estimation from CV data, the reliability of the adopted traffic models and assumptions are thus crucial. 
%However, if it were possible that the TSC system can solely use reliable data from observed traffic information source? Recent experiments on RL-TSC systems show such property and we will introduce them as following.

%\tcr{Learning (if capitalize first letter, be consistent; o.w. do NOT capitalize some but not others)}
\subsubsection{Reinforcement Learning (RL) based TSC with Partial CV Information}
\label{subsubsec:review_RL-based_TSC}

%\tcr{(seems we didn't highlight adaptive here. We may also need to emphasize that learning based TSC is primarily for adaptive control, not for other TSC control methods. Did PressLight stress that RL is primarily for adaptive?)} {wz: PressLight did not stress that. RL-TSC is actually known for adaptive TSC \citep{abdulhai2003reinforcement}; and most of papers on non-leanring based TSC using CV data are ATSC as well. But, yes, we can definitely stress that. I colored the changed text blue.}

%Comparing with non-learning-based TSC systems, RL-TSC methods show advantages in utilizing non-CV information during training. 

One major difference between non-learning based and RL based strategies lies in their dependency on traffic models. RL based methods do not rely on traffic models, and instead, they learn from past experience and adjust agents' behavior to local environments.

%\tco{separate the introduction of RL based TSC with CV by SARL and MARL. And strengthen the fact that the research under MARL-TSC with CVs are limited.}
{Recently, RL based multi-intersection TSC problem gains more attention from researchers \citep{chu2019multi,chacha2020toward,gong2019decentralized,yang2019cooperative}.} 
Despite an increasing number of papers on RL based TSC using CV data published in recent years \citep{kim2019real,hussain2020optimizing,yan2020efficiency,liu2014cooperative,liu2017distributed}, only a few pay attention to the performance of proposed RL based TSC systems under low CV penetration rate scenarios, {especially in large road networks.} \cite{aziz2019investigating} evaluate their previous RL based TSC method \citep{al2018minimizing} under various penetration rates in two real-world road networks. \cite{wu2020multi} also test the performance of their multi-agent RL based TSC algorithm under different penetration rates. In \cite{zhang2020using}, a series of more comprehensive experiments on the proposed RL based TSC method under different penetration rates and traffic demand patterns are conducted, { but those experiments are all limited in an isolated intersection}. Key features of these studies as well as the proposed method of this paper are summarized in Table \ref{tab:literature_review_rltsc}, including their traffic information source, environment, agent, algorithm, benchmarks, state, action, reward, and gap, respectively.

\cite{aziz2019investigating} find that their RL based TSC method can not learn well under penetration rates below 40\%; \cite{zhang2020using} and \cite{wu2020multi} show the robustness of proposed RL based TSC methods against low penetration rates. For instance, the RL based TSC system in \cite{zhang2020using} leads to an 80\% decrease in waiting time at the 20\% penetration rate, compared to its performance at 100\% penetration rate. The difference in the performance of these RL based TSC algorithms at low penetration rates can be explained by the design of RL based TSC systems and experiments. As mentioned before, \cite{zhang2020using} use non-CV delay as part of rewards during training, {but \cite{aziz2019investigating} did not}. Instead, CV data in \cite{aziz2019investigating} is utilized to estimate queue lengths in a simplified way: the distance between the position of the last stopping CV to the stop line is assumed to be the queue length of that lane. In addition, RL based TSC agents in \cite{aziz2019investigating} are trained at the 100\% penetration rate but tested under various penetration rates, which are not able to generalize well. As to \cite{wu2020multi}, a recurrent neural network (RNN) is applied, which can learn historic information from time-continuous traffic state data. Experiment results show that the system with an RNN layer is more robust against scenarios with partial observation available (20\% penetration rate or higher) than the same algorithm without such layer.

Previous studies in RL based TSC under various CV penetration rates demonstrate the great potential of RL based methods.
{However, an RL based TSC method using CV data and specifically designed for scenarios with low penetration rates and multiple intersections is missing.} It remains unclear what CV information can improve robustness against partial observability. Also, the RL structure suitable for partially observed systems remains to be explored.

%\tcr{lit. on multi-intersection RL and how we fill the gap?}

%\newpage

% \usepackage{booktabs}
% \usepackage{multirow}
\begin{landscape}
\begin{table}%[H]
\caption{Existing research of non-learning based TSC with low penetration rates}
\label{tab:literature_review_optimization_tsc}

\resizebox{\columnwidth}{!}{%
\begin{tabular}{@{}lllllllll@{}}
\toprule
References    & \begin{tabular}[c]{@{}l@{}}Information \\ Source\end{tabular}     & Environment    & \begin{tabular}[c]{@{}l@{}}Estimation \\ Granularity   \end{tabular}         & Estimation Methods                 & Benchmarks     & \multicolumn{1}{l}{\begin{tabular}[c]{@{}l@{}}Required minimum
%\tcr{need to explain the min. to reach what? o.w. the authors may not agree with the cut-off you picked here.}
\\ penetration rate$^1$ \end{tabular}} & Key Assumptions    & Gap       \\ \midrule
\cite{priemer2009decentralized}                                           & \begin{tabular}[c]{@{}l@{}}CV data; \\ Loop detectors\end{tabular}                                                          & \begin{tabular}[c]{@{}l@{}}A real-world network \\ of 9 intersections with \\ real-world traffic data\end{tabular}  & \begin{tabular}[c]{@{}l@{}}Flow-level \end{tabular} & \begin{tabular}[c]{@{}l@{}} Estimating queue length\\ based on current and \\historical CV data               \end{tabular}                                                                                                                                                                                                                                 & Well-tuned fixed-time TSC                                                                      & 33\%                                                                                             & \begin{tabular}[c]{@{}l@{}}No  constraints on \\the number of phases, \\phase  transitions, \\sequences or  timings    \end{tabular}                                                                                           &  \multirow{40}{*}{\begin{tabular}[c]{@{}l@{}} Non-learning based \\ methods usually rely \\on certain assumptions \\or traffic models  for\\ non-CV states estimation \end{tabular}}                                                                      \\ \cmidrule(r){1-8}
\cite{he2012pamscod}  & CV data                                                                                                                     & \begin{tabular}[c]{@{}l@{}}A real-world corridor \\ of 8 intersections\end{tabular} & \begin{tabular}[c]{@{}l@{}}Flow-level \end{tabular}  & \begin{tabular}[c]{@{}l@{}}Identifying platoons by analyzing \\ headways of CVs; estimating platoon\\ paratmeters using a linear regression\\ model.\end{tabular} & Well-tuned fixed-time TSC                                                         & 40\%                                                     & \begin{tabular}[c]{@{}l@{}}Penetration rate is given;\\ Passenger vehicles constitute \\ a significant majority of the \\ vehicles in the network\end{tabular} &     \\ \cmidrule(r){1-8}

\cite{lee2013cumulative}         & \begin{tabular}[c]{@{}l@{}}CV data; \\ Loop detectors \\ or camera\end{tabular}                                             & \begin{tabular}[c]{@{}l@{}}A synthetic isolated \\ intersection\end{tabular}                                        & \begin{tabular}[c]{@{}l@{}}Flow-level \end{tabular}  & \begin{tabular}[c]{@{}l@{}}Applying Kalman filtering \\ to estimate the cumulative \\ travel times\end{tabular}                                                                                                                                                                                                & Actuated TSC                                                                                   & 30\%                                                                                             & \begin{tabular}[c]{@{}l@{}}Total vehicle counts \\ are available\end{tabular}                                            &            \\\cmidrule(r){1-8}

\cite{goodall2013traffic}                                                                         & CV data                                                                                                                     & \begin{tabular}[c]{@{}l@{}}A real-world corridor \\ of 4 intersections\end{tabular}                                 & \begin{tabular}[c]{@{}l@{}}Flow-level \end{tabular} & \begin{tabular}[c]{@{}l@{}}Populating the CV data into a microscopic \\ traffic simulation to measure objective \\ function from the simulated behavior\end{tabular}                                                                                                                                           & Actuated TSC                                                                                   & 50\%                                                                                             & \begin{tabular}[c]{@{}l@{}}The turning movement\\  of vehicles are based\\  on their current lanes\end{tabular}                                                                  &                                                                        \\ \cmidrule(r){1-8}
\cite{goodall2014microscopic} & CV data                                                                                                                     & \begin{tabular}[c]{@{}l@{}}A real-world corridor \\ of 4 intersections\end{tabular}                                 & \begin{tabular}[c]{@{}l@{}}Vehicle-level \end{tabular} & \begin{tabular}[c]{@{}l@{}}Estimating the positions of\\ non-CVs based on the gaps \\among stopping CVs     \end{tabular}                                                                                  & \begin{tabular}[c]{@{}l@{}}The same TSC \\ without non-CV estimation\end{tabular}                  & \multicolumn{1}{l}{10\%-25\%}                                                                    & \begin{tabular}[c]{@{}l@{}}Vehicles behaviors follow \\ the Wiedemann car-following \\ model;\\ Length of CVs and gap between \\ vehicles are known\end{tabular}                 &                                                                        \\ \cmidrule(r){1-8}
\cite{tiaprasert2015queue}   & CV data                                                                                                                     & \begin{tabular}[c]{@{}l@{}}A synthetic isolated \\ intersection\end{tabular}        & \begin{tabular}[c]{@{}l@{}}Flow-level \end{tabular}  & \begin{tabular}[c]{@{}l@{}}Estimating queue length based on \\ stopping and moving CV data.\end{tabular}                                                           & \begin{tabular}[c]{@{}l@{}}Fixed-time TSC;\\ actuated TSC\end{tabular}            & Not Available                                                                         & \begin{tabular}[c]{@{}l@{}}Penetration rate is given;\\ The individual location and \\ speed of connected vehicle \\ can be collected\end{tabular}            &     \\ \cmidrule(r){1-8}

\cite{feng2015real,feng2016connected}                                                  & CV data                                                                                                                     & \begin{tabular}[c]{@{}l@{}}A real-world isolated\\ intersection    \end{tabular}                                                                                    & \begin{tabular}[c]{@{}l@{}}Vehicle-level \end{tabular}      & \begin{tabular}[c]{@{}l@{}}Using CV data to estimate speeds \\ and locations of non-CVs \end{tabular}                                                                                                                                                                                                   & Actuated TSC                                                                                   & \begin{tabular}[c]{@{}l@{}}25\% \citep{feng2015real},\\ $\leq 10\%$\citep{feng2016connected} \end{tabular}                                                                                           & \begin{tabular}[c]{@{}l@{}}Road segments of traffic \\ movements can be divided \\ into three regions;\end{tabular}                                                               &     \\ \cmidrule(r){1-8}
\cite{beak2017adaptive}  & CV data                                                                                                                     & \begin{tabular}[c]{@{}l@{}}A synthetic corridor\\of 5 intersections    \end{tabular}                                                                              & \begin{tabular}[c]{@{}l@{}}Vehicle-level \end{tabular} & \begin{tabular}[c]{@{}l@{}}Using CV data to estimate speeds \\ and locations of non-CVs\end{tabular}                                                                                          & Actuated TSC                                                                                   & \multicolumn{1}{l}{25\%}                                                              & \begin{tabular}[c]{@{}l@{}}Road segments of traffic \\ movements can be divided \\ into three regions\end{tabular} &         
    \\\cmidrule(r){1-8}
\cite{feng2018real}  & CV data                                                                                                                     & \begin{tabular}[c]{@{}l@{}}A real-world isolated\\ intersection    \end{tabular}                                                                              & \begin{tabular}[c]{@{}l@{}}Flow-level \end{tabular} & \begin{tabular}[c]{@{}l@{}}Estimating cycle-by-cycle vehicle\\  arrival times and delays based \\ on estimated average historical \\ volume and a limited number of \\ observed critical CV trajectories\end{tabular}                                                                                          & Actuated TSC                                                                                   & \multicolumn{1}{l}{\textless{}10\%}                                                              & \begin{tabular}[c]{@{}l@{}}Vehicle arrivals follow \\ Poisson process;\\ Vehicle length is uniform \\ and known:\\ Free travel times are the \\ same for all vehicles\end{tabular} &         
    \\\cmidrule(r){1-8}
\cite{mohebifard2019cooperative} & \begin{tabular}[c]{@{}l@{}}CV data; \\ Loop detectors;\\ Infrastructure to \\ infrastructure \\ communications\end{tabular} & \begin{tabular}[c]{@{}l@{}}A real-world network \\ of 20 intersections\end{tabular} & \begin{tabular}[c]{@{}l@{}}Flow-level \end{tabular}  & \begin{tabular}[c]{@{}l@{}}Estimating desnity across network \\ links using data from CVs and loop detectors\end{tabular}                                                & \begin{tabular}[c]{@{}l@{}}The solution \\ to the central \\ problem\end{tabular} & 30\%                                                     & \begin{tabular}[c]{@{}l@{}}The dynamic travel demand \\ is known;\\ vehicle counts and stop bar\\ detectors are available at\\ all intersections\end{tabular} &     \\\cmidrule(r){1-8}

\cite{al2020real}                 & \begin{tabular}[c]{@{}l@{}}CV data; \\ Loop detectors;\\ Infrastructure to \\ infrastructure \\ communications\end{tabular} & \begin{tabular}[c]{@{}l@{}}A real-world corridor \\ of 4 intersections with \\ real-world traffic data\end{tabular} & \begin{tabular}[c]{@{}l@{}}Flow-level \end{tabular}     & \begin{tabular}[c]{@{}l@{}}1) Integrating data from CVs \\ and loop detectors to estimate the \\ trajectories of non-CVs \\ based on car-following concepts. \\ 2) Converting the temporal point vehicle \\ detections to a spatial vehicle \\ distribution on a link.\end{tabular} & \begin{tabular}[c]{@{}l@{}}Real-world signal plan \\ in the case study; \\ Vistro\end{tabular} & 0\%                                                                                              & Stop bar detectors are available                                                                                                                                                  &  \\ \bottomrule
\end{tabular}
}
\begin{tablenotes}
      \item 1: The ``required minimum penetration rate'' refers to the minimum penetration rate where the proposed method can outperform the performance of its benchmarks in the experimental environment. For studies that do not provide with such information, we use ``Not Available'' instead.
\end{tablenotes}
\end{table}
\end{landscape}

% \usepackage{booktabs}
\begin{landscape}
\begin{table}%[h]
\caption{Existing research of RL based TSC with low penetration rates}
\label{tab:literature_review_rltsc}
\resizebox{\columnwidth}{!}{%
\begin{tabular}{@{}llllllllll@{}}
\toprule
Reference  & \begin{tabular}[c]{@{}l@{}} Information \\ Source\end{tabular}  & Environment  & Agent  & Algorithm & Benchmarks   & State  & Action & Reward  & Gap                         \\ \midrule
\cite{aziz2019investigating}    & CVs                        & \begin{tabular}[c]{@{}l@{}}A real-world \\ 4-intersection \\ corridor; \\ A real-world \\ network of 20 \\ intersections\end{tabular}   & \begin{tabular}[c]{@{}l@{}} TSC of one\\ intersection\end{tabular} & Q-learning       & \begin{tabular}[c]{@{}l@{}} The proposed \\ algorithm with \\different rewards \end{tabular}& \begin{tabular}[c]{@{}l@{}} Estimated queue lengths\end{tabular} & \begin{tabular}[c]{@{}l@{}}Keep current phase \\or change to the\\ next phase \end{tabular} & \begin{tabular}[c]{@{}l@{}} Delay;\\Energy consumption;\\Or energy consumption\\ with penalty for stops \end{tabular}  &   \multirow{16}{*}{ \begin{tabular}[c]{@{}l@{}} (1) An RL based TSC system using CV data \\and specifically designed for scenarios \\with low penetration rates is missing;\\ (2) It remains unclear what \\CV information could improve\\ robustness against partial \\observation from CVs;\\ (3) The RL structure suitable for\\ partially observed system remains\\ to be explored. \end{tabular} } \\\cmidrule(r){1-9}
\cite{wu2020multi} & CVs, camera                & \begin{tabular}[c]{@{}l@{}} A 2-3 synthetic$^1$ \\ network\end{tabular}     & \begin{tabular}[c]{@{}l@{}} TSC of one\\ intersection\end{tabular}   & MARDDPG   & \begin{tabular}[c]{@{}l@{}}Fixed-time;\\ DDPG; \\SOTL;\\ IDQN \end{tabular}                                & 
\begin{tabular}[c]{@{}l@{}} Vehicle location;\\ Vehicle velocity;\\ Queue length;\\ Current traffic \\ light phase; \\Number of pedestrians  \end{tabular}  & \begin{tabular}[c]{@{}l@{}}Keep current phase \\or change to the\\ next phase \end{tabular}  & \begin{tabular}[c]{@{}l@{}} Weighted sum of delay; \\Queue length;\\ Throughput;\\ Blinking condition \\of traffic;\\ Waiting time \\of vehicles \\and pedestrians \end{tabular}  &  \\ \cmidrule(r){1-9}

\cite{zhang2020using}    & CVs                        & \begin{tabular}[c]{@{}l@{}}A synthetic isolated \\ intersection \end{tabular}   & \begin{tabular}[c]{@{}l@{}} TSC of one\\ intersection\end{tabular} & DQN       & \begin{tabular}[c]{@{}l@{}}Agent trained and \\tested under the \\same scenario;\\ Fixed-time TSC \end{tabular}& \begin{tabular}[c]{@{}l@{}}The distance to the \\nearest detected vehicle;\\ Number of detected vehicles;\\ Amber phase indicator;\\ Current traffic light \\Current phase elapsed time;\\ Current time of a day;\\ Current phase\end{tabular} & \begin{tabular}[c]{@{}l@{}}Keep current phase \\or change to the\\ next phase \end{tabular} & \begin{tabular}[c]{@{}l@{}} Negative value of\\ speed loss \end{tabular}  &   \\\hline
%\cmidrule(r){1-9}
CVLight (this paper)    & CVs                        & \begin{tabular}[c]{@{}l@{}}A 2-2 synthetic \\network; \\ A 5-5 synthetic \\network; \\ A 2-2 real-world \\ network\end{tabular}   & \begin{tabular}[c]{@{}l@{}} TSC of one\\ intersection\end{tabular} & Asym-A2C$^2$     & \begin{tabular}[c]{@{}l@{}} Max-Pressure\\PressLight\\DQN\\Webster's method\\Actuated TSC \end{tabular}& \begin{tabular}[c]{@{}l@{}} Number of vehicles in each \\incoming and outgoing lane;\\Average delay per vehicle \\of each incoming lane;\\Current traffic signal phase;\\Current phase elapsed time\\\end{tabular} & \begin{tabular}[c]{@{}l@{}}Choose the \\next phase \end{tabular} & \begin{tabular}[c]{@{}l@{}}Negative value of \\ pressure on \\the intersection \end{tabular}  &   \\\bottomrule
\end{tabular}
}
\end{table}
\begin{tablenotes}
      \item 1. The ``2-3'' represents the 2-by-3 road network structure.
      \item 2. We will introduce the Asym-A2C in Section \ref{subsec:a2c_method}.
\end{tablenotes}
\end{landscape}

% add one row of our proposed method and show our advantages

\subsection{Contributions of This Paper}
\label{subsec:contribution}

{To mitigate the limits we mentioned in the above section, this paper proposes a reinforcement learning scheme, CVLight, for the multi-intersection TSC problem without a full CV penetration rate. To investigate what CV information can improve robustness against partial observation, we incorporate vehicle delays and phase durations into the state design. By exploring the structure of the actor-critic algorithm, we include the information from both CVs and non-CVs into training. }

{Specifically, the main contributions of this paper are listed below}:

{
\begin{enumerate}
    %\item \tcr{develop a reinforcement learning scheme, CVLight, (not specific, there are a considerable amount of lit. on RL for TSC. What's unique about ours? where u highlight multi-agent?)} and validate the scheme by comparing it with state-of-the-art benchmark algorithms and visualizing the learned policies.  
    \item {We model the TSC system that leverages CVs information using a partial observable reinforcement learning scheme, denoted as CVLight. We design states and rewards for CVLight to foster inter-agent coordination and consider delays and phase durations.}
    \item {We develop a novel training algorithm, Asymmetric Advantage Actor Critic (Asym-A2C), which utilizes asymmetric information: the actor is trained with the partial observation of CVs while the critic is trained with the full observation of both CVs and non-CVs.} 
    \item {We demonstrate the advantage of CVLight over benchmarks using a comprehensive list of numerical experiments. We also demonstrate that, by utilizing pre-trained models of small road networks, CVLight can be applied to larger road networks at lower computation costs.}
    %\item exploit the architecture of the advantage actor-critic (A2C) algorithm by utilizing both CV and non-CV data in training and only CV in execution, which can help stabilize training for low CV penetration. 
    %\item validate the performance of CVLight specifically for a real-world scenario on a network of four intersections with various CV penetration rates. 
\end{enumerate}
}
%\tcr{development of a RL based TSC system to solve traffic congestion in an environment with low penetration rates. (this statement is vague, still unclear what the contribution is. Many others including Jeff Ban also proposed RL based TSC.)} Inspired by the centralized training and decentralized execution paradigm, the proposed A2C structure allow critics to receive the true traffic information from CVs and non-CVs during the training stage while relying on actors to choose actions based on CV data only in the execution stage. 
% \tcr{discussion (discussion may not be deemed as contribution.)} of what information is critical for RL based TSC agent learning under low penetration rate scenario. \tcr{your tone is too modest. From your statement, I don't feel this paper contains sufficient contributions. It mainly discusses things instead of making progresses in this field.}\item design states and rewards for better training performance

The remainder of this paper is structured as follows. {Section \ref{sec:algorithm} presents the basics of reinforcement learning in the context of TSC with CVs and outlines the CVLight model. The proposed model is then examined in Section \ref{sec:experiments} to validate the design, including performance comparison, sensitivity analysis, and scalability in road networks sizes. In Section \ref{sec:case_study}, CVLight is compared to multiple benchmarks on real-world intersections. Finally, we conclude the paper and present future research directions in Section~\ref{sec:discussions&conclusion}.}

\section{Reinforcement Learning based Traffic Signal Control (RL-TSC)}
\label{sec:algorithm}

In this section, we will introduce concepts and terminologies in reinforcement learning based traffic signal control (RL-TSC), taking one state-of-the-art RL-TSC algorithm known as PressLight \citep{wei2019presslight} as an example. Based on this, we discuss the limitation of Presslight under scenarios without a full penetration rate and propose a new model, \emph{CVLight}.

%Based on this, we will introduce one state-of-the-art RL-TSC algorithm, known as PressLight \citep{wei2019presslight}, and discuss its limitation under partial\tco{(low)} penetration rate scenarios through a simple example \tco{(actually we don't have such example.)(wz: we introduce the )}. To address this limitation, we will develop\tco{(propose, delete "will")} a new model, \emph{CVLight}, in this section.
%\tcr{there is a confusion between "CVLight" and "A2C-Full". Both are referred to as algorithm, which is confusing. Are you referring "A2C-Full-Sep" as "CVLight"?} wz: CVLight actually refers to the specific model, A2C-psr-duration (Full-Sep), i.e. the A2C-psr-duration design solved by A2C-Full-Sep algorithm.

Prior to delving into the introduction of RL and our proposed model, we first define major notations that will be used in the subsequent sections.
%\tco{(I don't think a space is "a set". Maybe we can just delete "a set of" )}

\nomenclature[A, 01]{$\mathbf{s}$}{state}
\nomenclature[A, 02]{$S$}{state space}
\nomenclature[A, 03]{$a$}{action}
\nomenclature[A, 04]{$A$}{action space}
\nomenclature[A, 04.1]{$o$}{observation}
\nomenclature[A, 05]{$O$}{observation space}
\nomenclature[A, 05.5]{$r$}{reward}
\nomenclature[A, 06]{$R$}{reward space }
\nomenclature[A, 06.1]{$Pr$}{state transition function}
\nomenclature[A, 06.3]{$\gamma$}{ discounted factor}
\nomenclature[A, 08]{$\pi$}{policy}
\nomenclature[A, 09.4]{$Q(s,a)$}{state-action value function}
\nomenclature[A, 10]{$V(s)$}{state value function}
\nomenclature[A, 11]{$\theta$}{parameters of critic's value neural network}
\nomenclature[A, 12]{$\phi$}{parameters of actor's policy neural network}

\nomenclature[B,01]{$I$}{set of all intersections}
\nomenclature[B,02]{$L_{in}(i)$}{ set of incoming lanes of intersection $i$}
\nomenclature[B,02.5]{$L_{out}(i)$}{set of outgoing lanes of intersection $i$}
\nomenclature[B,04]{$p_i$}{current phase of intersection $i$}
\nomenclature[B,05]{$d_i$}{current phase duration of intersection $i$}
\nomenclature[B,06]{$n_i(l)$}{number of vehicles in lane $l$ at intersection $i$}
\nomenclature[B,07]{$x_i(l)$}{scaled average delay per vehicle of lane $l$ at intersection $i$}
\nomenclature[B,08]{$t_{min}(l)$}{expected minimum travel time of lane $l$}
\nomenclature[B,09]{$D_i$}{cumulative delay of all vehicles at intersection $i$}
\nomenclature[B,10]{$P_i$}{pressure of intersection $i$}
\printnomenclature

%\tcr{pls rephrase this section. Many parts including A2C sound very like Zhenyu's paper.} {Sure. I will re-write this section and make new illustrations to introduce how does our RL-TSC system interact with an environment with low penetration rates.}

\subsection{Preliminaries}
\label{subsec:preliminaries}

For an isolated intersection, we regard the control unit of traffic signals as an \emph{agent}, and all other things as \emph{environment}. The TSC problem can be formulated as a Markov decision process (MDP), which is specified by a tuple of $(S,A,R,Pr,\gamma)$. The state space $S$ contains all necessary information describing the environment, such as the queue length of each lane. $A$ is a set of actions that an agent can perform to interact with the environment, e.g. to determine the length of the next traffic signal phase. Given a certain state $s \in S$, the agent chooses one action $a \in A$ following its policy $\pi(a|s)$, which maps from the state to actions in the action space $A$.

\begin{figure}[H]
\centering
\includegraphics[scale=0.4]{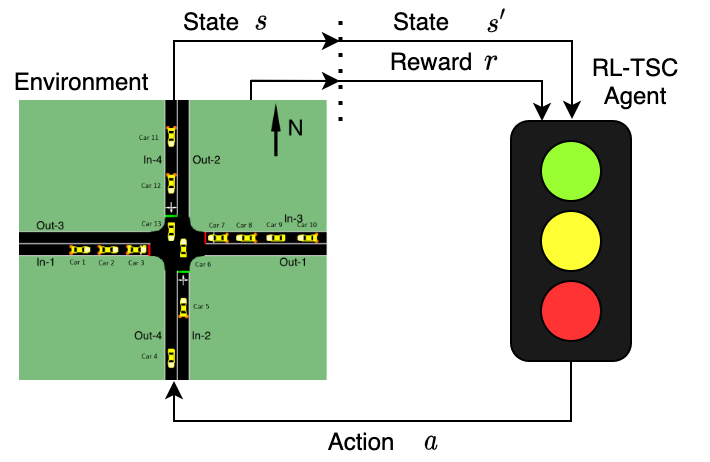}
\caption{An illustration of interaction between RL-TSC agent and traffic environment}
\label{fig:ill-rl-tsc}
\end{figure}

The state transition function $Pr$ is a perfect model of the environment's mechanism. As shown in Figure \ref{fig:ill-rl-tsc}, once the newly chosen action is executed, the agent can observe a new state $s'$ based on $Pr$ and receive a reward $r \in R$. The goal of RL is to find the optimal policy $\pi^*$ that can maximize the return, which is a function of a sequence of rewards weighted by a discounted factor $\gamma$. To estimate the expected return under a certain circumstance, two kinds of value functions are proposed: a state value function $V^\pi (s)$ estimates how favorable it is to be in the state $s$ under policy $\pi$; and a state-action value function $Q^\pi (s,a)$ estimates how favorable it is to take action $a$ from state $s$ under policy $\pi$. For a complicated real-world traffic environment, the state transition function $Pr$ is unknown. In this case, the MDP problem can be solved by Temporal Difference (TD) techniques, such as Q-learning and Policy Gradients. More details about RL can be found in \cite{sutton1998introduction}. 
%tco{(Delete tis paragraph. wz: you mean combine this one and the above one?)}

To better illustrate how an RL-TSC problem is formulated, we use PressLight \citep{wei2019presslight} as an example. Theoretically grounded by a popular TSC algorithm, Maxpressure \citep{varaiya2013max}, PressLight is proven to outperform Maxpressure and other recent RL-TSC algorithms, such as GRL \citep{van2016coordinated}, under multi-intersections control scenarios.

\begin{figure}[H]
\centering
\includegraphics[scale=0.3]{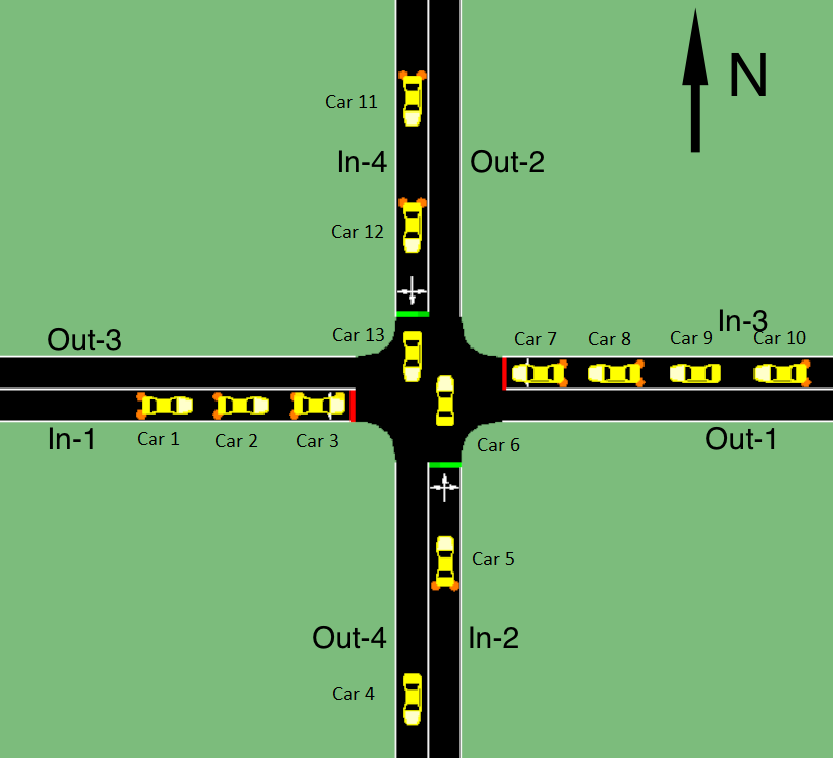}
\caption{An example of state for PressLight}
\label{fig:rltsc-example}
\end{figure}

%The traffic state of the example is shown in Figure \ref{fig:rltsc-example}.
%\tcr{should define state, action, reward in bullet points clearly and formulate single-agent RL first.}

% would it be too complicated if we use PressLight as the example here? 
The state of PressLight is defined as a vector of the current traffic signal phase and the number of vehicles in each incoming and outgoing lane. For the intersection shown in Figure \ref{fig:rltsc-example}(a), the state vector can be represented as: 
%$s = [0,3,1,4,2,0,0,0,1]$. 
\begin{equation*}
    s = [\overbrace{(1,0)}^{\text{phase}},\underbrace{(3,1,4,2)}_{\text{number of vehicles }\atop\text{in incoming lanes}},\overbrace{(0,0,0,1)}^{\text{number of vehicles}\atop\text{in outgoing lanes}}],
\end{equation*}

%\parbox[position]{width}{text}

\noindent where the first entry %\tco{(I suggest you to use () for each entry, like: [(1,0), (3,1,4,2), (0,0,0,1)])} 
of $s$, {a one-hot vector}, represents the current green phase for the North-South direction (denoted by NS-allow phase); the vector $(3,1,4,2)$ refers to the numbers of vehicles in incoming lanes from incoming lane 1 (denoted by In-1) to In-4, respectively; similarly, each entry of the vector $(0,0,0,1)$ represents the number of vehicles in the corresponding outgoing lane, from Out-1 to Out-4 (Car 6 and 13 are not counted as they are not in any incoming or ongoing lane). 
%\tcr{(such a long vector with some random numbers is not easy to follow from the beginning. I have to patiently wait to finish reading the entire paragraph. suggest indicating what each component represents underneath using arrows and texts upfront, then you can even skip cumbersome description below.)}
%The traffic signal controller has two phases, the green phase for North-South direction (denoted by NS-allow and encoded as 0) or for East-West direction (EW-allow, encoded as 1). Currently, the phase is encoded as 0, which is the first entry of $s$. The number of vehicles in each incoming lane can be represented as a vector such as [3,1,4,2] in the Figure \ref{fig:rltsc-example} (a), where each number in the vector corresponds to the number of vehicles in one lane, starting from incoming lane 1 (denoted by In-1) to In-4. For example, the entry of the first index, i.e. 3, corresponds to car 1, 2, and 3 on In-1. In the same manner, the number of vehicles on each outgoing lane can be represented as a vector such as [0,0,0,1]. The entry of the fourth index, 1, represents the number of vehicles on the outgoing lane 4 (Out-4) (Car 6 and 13 are not counted as they are not on any ongoing lanes). 

%{wz: add the description about decision point and detail the action as follows.} 
%\tcr{You presented a figure to demonstrate our CVLight in terms of time decision points. Can we add it when CVLight is introduced?}
Observing the current state $s$, the PressLight agent chooses its action $a$. The action is an index of one signal phase that will be executed as the next phase. If the index is the same as the index of the current phase, the current phase will continue for a pre-defined length of time, e.g. 7 seconds; if not, TSC will first execute a yellow phase and an all-red clearance phase consecutively for a pre-defined length of time and then execute the chosen traffic signal phase for the minimum green time. 
%\tcr{what is the decision time point? How long will the phase be? The action description is not clear. If necessary, can draw a time arrow and indicate decision points and state. I remembered you guys have some nice TSC phase figures that may be more informative.} 

After the execution of action $a$, the agent can observe a new state $s'$ and receive a reward $r$, which indicates how good its chosen action is. For PressLight, the reward is calculated based on the current \emph{pressure}, which is defined as the sum of the difference between the numbers of vehicles in incoming lanes and outgoing lanes of all phases and is presented in Equation \ref{eqa:pressure}. For simplicity, we use the traffic state $s$ presented in Figure \ref{fig:rltsc-example}(a) as an example for pressure calculation. Assuming no turning traffic flows, the pressure of the NS-allow phase is the sum of the difference between the number of vehicles in In-2 and Out-2 and the difference between the number of vehicles in In-4 and Out-4. Thus, the pressure of NS-allow phase is 3. Similarly, the pressure of EW-allow phase is 7 and the total pressure of this intersection is 10 (here we do not consider the lane capacity in Equation \ref{eqa:pressure}).

\subsection{{CVLight for a Single Intersection.}} 
\label{subsec:agent_design_sarl}

\begin{figure}[H]
  \subfloat[State (all cars are observable)]{
	\begin{minipage}[c][0.95\width]{
	   0.5\textwidth}
	   \centering
	   \includegraphics[width=1\textwidth]{RL-TSC-example.png}
	\end{minipage}}
 \hfill 	
  \subfloat[Observation (only red cars are CVs)]{
	\begin{minipage}[c][0.95\width]{
	   0.5\textwidth}
	   \centering
	   \includegraphics[width=1.0\textwidth]{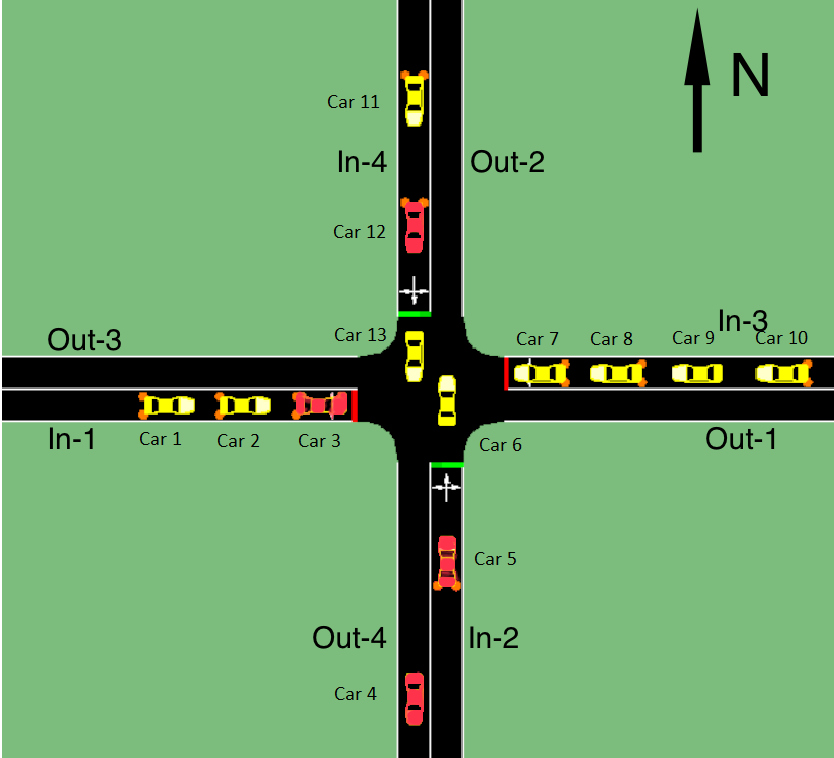}
	\end{minipage}}

\caption{An example of state and observation for CVLight (Note that In-1, for example, represents the incoming lane 1 and Out-1 represents the outgoing lane 1)}
\label{fig:cvlight-sarl-example}
\end{figure}

{The previous example in Section \ref{subsec:preliminaries} is a simple scenario where the RL agent can obtain complete information about the environment.} 
However, it might not work in the CV setting. %For example, in a multi-intersection scenario, if each traffic signal controller of an intersection is modeled as an agent and each agent can only detect vehicle information of its local intersection, 
When only partial traffic information is known, the problem becomes a Partially Observable Markov Decision Process (POMDP). In POMDP, agents can not obtain complete information about the environment's state; instead, they can only obtain partial information, denoted by observation $o \in O$, where $O$ is the observation space. %Besides multi-agent RL-TSC, for a RL-TSC system using CV data, if the penetration rate is not 100\% and non-CVs information is not available, this problem can also be modeled as a POMDP and the observation is mainly composed of the traffic states of CVs.
%As to the POMDP for the multi-agent RL problem, PressLight is proven to be effective \citep{wei2019presslight}. But for the POMDP under partial connectivity, algorithms such as 
It is not appropriate to directly apply PressLight to scenarios with CV penetration rates below 100\%. %, because the original state and reward design might misguide the agent. 
For example, in Figure \ref{fig:cvlight-sarl-example}(a), if only four vehicles are CVs and the number of vehicles in observation can only be measured based on CVs, then the observation $o$ can be significantly different from the state $s$. {Under such a circumstance, the PressLight agent is not able to detect the congestion on In-1 and In-3 and may learn wrong policies due to the significant difference between states and observations.}

{To cope with the issues mentioned above, we propose \emph{CVLight} and introduce its state, action, and reward design in the single-intersection TSC problem. }

%\tcr{This section needs to be rewritten. We need to clearly highlight the proposed model, which should be prs-duration, and directly frame it on a corridor with multiple intersections and independent learner. All other models are baselines.}

%{We will introduce the state and reward design of our proposed model, \emph{CVLight}. Below we will describe the agent design in a multi-intersection scenario, in which an agent, indexed by $i$, represents a traffic signal controller at one intersection and owns its neural networks.}

%{Reviewer\#1 1. The paper assumes that CVs use either V2V or V2I. Please clarify what information CVs provide, and how different communication mechanisms (V2I or V2V) would impact the application of the algorithm.}

%{Reviewer\#1 2. In the representation of states, the signal phase is a categorical variable, which is typically modeled through a set of binary variables. However, it seems that the representation in the paper treats the signal phase as an integer number. Please justify such a choice.}

1) \textbf{State}.  The state $s$ is defined as a vector of current phase $p$, elapsed time of current phase (named as phase duration $d$), number of vehicles in each incoming lane $n(l)$ ($l \in L_{in}$, $L_{in}$ is a set of incoming lanes), number of vehicles in each ongoing lane $n(m)$ ($m \in L_{out}$, $L_{out}$ is a set of outgoing lanes), and {delay ratio $x(l)$ ($l \in L_{in}$), which is the ratio of the average delay per vehicle of each incoming lane and the total minimum travel time of all vehicles in the incoming lanes}. In summary, the state $s$ is written as below:

\begin{equation}
    %s_i = [\underbrace{p_i}_{\text{phase}},\underbrace{d_{i}}_{\text{duration}},\underbrace{n_i(l),n_i(m)}_{\text{number of vehicles in lane}},\underbrace{x_i(l)}_{\text{average delay}}], for~l\in L_{in}(i), m \in L_{out}(i).
    s = [\overbrace{p}^{phase},\underbrace{d}_{\text{phase duration}},\overbrace{n(l)}^{\text{number of vehicles}\atop\text{in incoming lanes}},\underbrace{n(m)}_{\text{number of vehicles}\atop\text{in outgoing lanes}},\overbrace{x(l)}^{\text{average delay per}\atop\text{vehicle in incoming lanes}}], ~for~l\in L_{in}, m \in L_{out}.
\end{equation}

Below we will describe how the value of each component is obtained. 

\begin{itemize}
    \item {The phase $p$ is a one-hot vector that indicates the current traffic phase at the intersection.}
    
    \item Due to the limited available information from CVs, if the RL agent can not observe CVs in one direction, it is possible that the RL agent maintains the green light in another direction for a very long time, which is unrealistic and unfair. Considering this, we include phase duration $d$ into the state, which enables the agent to build a connection between the phase duration and the observed delay of vehicles.
 
    \item Similar to PressLight \citep{wei2019presslight} and Maxpressure \citep{varaiya2013max}, we include the number of vehicles in each incoming and outgoing lane as part of the state. Specifically, $n(l)$ is the vehicle count in a certain lane within a given detected radius. In this paper, the detected radius is 200 meters. If the lane length is shorter than this default radius, the radius will be adjusted to the actual lane length. { We divide the road into three segments for each lane so as to provide spatial distribution information of vehicles with the agent \citep{yang2018queue,wei2019presslight}.}
    
    %{Reviewer\#1 3. It is claimed in the paper that "incoming lane is not separated into segments because our initial experiments show no significant improvement under our experimental environment." Considering the spatial distribution of vehicles can improve the estimation accuracy of queue length, and therefore improve the performance of the signal controller [1-2], so this conclusion seems to be quite counter-intuitive. Please provide more insights on this, especially whether and how the A2C framework learns such a mechanism internally. [1] Hao, P., & Ban, X. (2015). Long queue estimation for signalized intersections using mobile data. Transportation Research Part B: Methodological, 82, 54-73. [2] Yang, K., & Menendez, M. (2018). Queue estimation in a connected vehicle environment: A convex approach. IEEE Transactions on Intelligent Transportation Systems, 20(7), 2480-2496.}
    
    \item Travel delay of a vehicle is defined as the difference between the current travel time and the expected minimum travel time of lane $l$, denoted by $t_{min}(l)$. The minimum travel time of all vehicles in that lane and is defined as

\begin{equation}
        t_{min}(l) = \frac{length(l)}{v_{max}(l)},
\end{equation}

\noindent where 
\begin{itemize}
    \item $length(l)$ is the length of lane $l$;
    \item $v_{max}(l)$ is the speed limit of lane $l$.
\end{itemize}

Therefore, the scaled average delay per vehicle in incoming lane $l$,  $x(l)$, is defined as

\begin{equation}
        x(l) = \frac{\sum\limits_{v \in V(l, t) } \left( t - t_0(v, l) - t_{min}(l) \right)}{n(l)~ t_{min}(l)}, l \in L_{in},
\end{equation}

\noindent where 
\begin{itemize}
    \item[--] $V(l,t)$ is the set of all vehicles in lane $l$ at time step $t$;
    \item[--] $t$ is the current time step;
    \item[--] $t_0(v, l)$ is the time step when vehicle $v$ enters lane $l$.
\end{itemize}

Since the delay is cumulative over time, it would contain more historical information than the number of vehicles in a lane, even with limited CV data available. Thus, we add the average delay per vehicle into our state as well to better utilize the CV information, especially for low penetration rate scenarios. 
    
%{Reviewer\#1 (minor) 1. By Eq.(2.2), the variable $x_i$ is not the delay, but more a delay ratio, since the unit is not in seconds. }

%\tcr{How about other components explanation?}
\end{itemize}

2) \textbf{Observation}. {We assume that CVs will send their real-time locations and IDs using V2I technologies to both the upstream and downstream intersections. Partial information that is observed by CVs is denoted below (note that we include the traffic signal information, $p$ and $d$, into $o_{CV}$ as well):}
\begin{equation}
    %s_i = [\underbrace{p_i}_{\text{phase}},\underbrace{d_{i}}_{\text{duration}},\underbrace{n_i(l),n_i(m)}_{\text{number of vehicles in lane}},\underbrace{x_i(l)}_{\text{average delay}}], for~l\in L_{in}(i), m \in L_{out}(i).
    o_{CV} = [\overbrace{p}^{phase},\underbrace{d}_{\text{phase duration}},\overbrace{n_{CV}(l)}^{\text{number of CVs}\atop\text{in incoming lanes}},\underbrace{n_{CV}(m)}_{\text{number of CVs}\atop\text{in outgoing lanes}},\overbrace{x_{CV}(l)}^{\text{average delay per}\atop\text{CV in incoming lanes}}], ~for~l\in L_{in}, m \in L_{out}.
\end{equation}

{Partial information of non-CVs (denoted by $o_{\overline{CV}}$), which is not observed by CVs, is denoted as:}

\begin{equation}
    %s_i = [\underbrace{p_i}_{\text{phase}},\underbrace{d_{i}}_{\text{duration}},\underbrace{n_i(l),n_i(m)}_{\text{number of vehicles in lane}},\underbrace{x_i(l)}_{\text{average delay}}], for~l\in L_{in}(i), m \in L_{out}(i).
    o_{\overline{CV}} = [\overbrace{n_{\overline{CV}}(l)}^{\text{number of non-CVs}\atop\text{in incoming lanes}},\underbrace{n_{\overline{CV}}(m)}_{\text{number of non-CVs}\atop\text{in outgoing lanes}},\overbrace{x_{\overline{CV}}(l)}^{\text{average delay per}\atop\text{non-CV in incoming lanes}}], ~for~l\in L_{in}, m \in L_{out}.
\end{equation}

2) \textbf{Action}. Traffic signal phases are acyclic. Once the RL agent observes $o_{CV}$, it can select one phase $p$ as its action $a$ from its phase set (action set) $A$. If $p$ is the same as the current phase, the current phase will continue for a certain period of time (in this paper, we use the minimum green time, which is 7 seconds); or if $p$ is different from the current phase, the intersection will first consecutively experience a yellow phase (1 second) and an all-red clearance phase (2 seconds), and then execute the chosen phase $p$ for a minimum green time. To avoid extreme behavior such as keeping one phase for a long time, we enforce the maximum green time (40 seconds by default) for phase duration. {Once the current phase duration exceeds the maximum green time, the agent will be enforced to choose the phase with the highest probability apart from the current phase as the next phase.}

%{Reviewer\#1 4. "Once the current phase duration overpasses the maximum green time, the agent will randomly choose one of phases other than the current phase as the next phase." Please clarify why to choose a random phase rather than optimizing the phase as well.}

%{Reviewer\#2 9. The control interval is set as 5 sec. If the penetration rate is low, then the agent selects actions randomly very often (due to scarce data from CVs). Given this, what are the point of choosing a small control interval, and what performance the CVLight could deliver under different (longer) control intervals?}

3) \textbf{Reward}. {The reward $r$ adopts the reward from PressLight\citep{wei2019presslight}, which is the negative value of the pressure:} 
%\tco{I think it might be better to cite MaxPressure paper instead of PressLight. Wz: the definition of Pressure in MaxPressure is different from here. We adopt the form from PressLight, which normalizes the pressure with the lane capacity.}

\begin{equation}\label{eqa:pressure}
    r = - P = - \left| \sum_{(l,m)\in L_{phase}} \left( \frac{n(l)}{n_{max}(l)} - \frac{n(m)}{n_{max}(m)} \right) \right|,
\end{equation}

\noindent where 
\begin{itemize}
    \item[--] $P$ is the pressure;
    \item[--] $L_{phase}$ is a set of combinations of incoming lane $l \in L_{in}$ and outgoing lane $m\in L_{out}$ of each traffic movement controlled by traffic signal phases; 
    \item[--] $n(l)$ is the number of vehicles in lane $l$; 
    \item[--] $n_{max}(l)$ is the capacity of lane $l$, i.e. the maximum number of vehicles the lane can store.
\end{itemize}

\subsection{{CVLight for Multiple Intersections}}
\label{subsec:agent_design_marl}

%\tco{(introduce the difference of equations under MARL; introduce how pressure helps coordination under MARL.)}

{We will introduce the agent design of \emph{CVLight} for scenarios of multiple intersections}. %\st{Below we will describe the agent design in a multi-intersection scenario, in which an agent, indexed by $i$, represents a traffic signal controller at one intersection and owns its neural networks.}}

%{Reviewer\#1 1. The paper assumes that CVs use either V2V or V2I. Please clarify what information CVs provide, and how different communication mechanisms (V2I or V2V) would impact the application of the algorithm.}

%{Reviewer\#1 2. In the representation of states, the signal phase is a categorical variable, which is typically modeled through a set of binary variables. However, it seems that the representation in the paper treats the signal phase as an integer number. Please justify such a choice.}

1) \textbf{State}. 
\begin{equation}
    %s_i = [\underbrace{p_i}_{\text{phase}},\underbrace{d_{i}}_{\text{duration}},\underbrace{n_i(l),n_i(m)}_{\text{number of vehicles in lane}},\underbrace{x_i(l)}_{\text{average delay}}], for~l\in L_{in}(i), m \in L_{out}(i).
    s_{i} = [\overbrace{p_{i}}^{phase},\underbrace{d_{i}}_{\text{phase duration}},\overbrace{n_{i}(l)}^{\text{number of vehicles}\atop\text{in incoming lanes}},\underbrace{n_{i}(m)}_{\text{number of vehicles}\atop\text{in outgoing lanes}},\overbrace{x_{i}(l)}^{\text{average delay per}\atop\text{vehicle in incoming lanes}}], ~for~l\in L_{in}(i), m \in L_{out}(i),
\end{equation}
where the subscript is the index of each intersection. $L_{in}(i)$ and $L_{out}(i)$ are sets of incoming and outgoing lanes of the intersection $i$, respectively. 

{2) \textbf{Observation.}} 

\begin{equation}
    o_{i,CV} = [\overbrace{p_i}^{phase},\underbrace{d_i}_{\text{phase duration}},\overbrace{n_{i,CV}(l)}^{\text{number of CVs}\atop\text{in incoming lanes}},\underbrace{n_{i,CV}(m)}_{\text{number of CVs}\atop\text{in outgoing lanes}},\overbrace{x_{i,CV}(l)}^{\text{average delay per}\atop\text{CV in incoming lanes}}], ~for~l\in L_{in}(i), m \in L_{out}(i).
\end{equation}

\begin{equation}
    o_{i,\overline{CV}} = [\overbrace{n_{i,\overline{CV}}(l)}^{\text{number of non-CV}\atop\text{in incoming lanes}},\underbrace{n_{i,\overline{CV}}(m)}_{\text{number of non-CVs}\atop\text{in outgoing lanes}},\overbrace{x_{i,\overline{CV}}(l)}^{\text{average delay per}\atop\text{non-CV in incoming lanes}}], ~for~l\in L_{in}(i), m \in L_{out}(i).
\end{equation}

2) \textbf{Action}. Intersection $i$ chooses action $a_{i}$ independently, once it observes the $o_{i,CV}$. {Note that agents take actions asynchronously, and thus the time steps when agents make new actions can be different from each other.}

3) \textbf{Reward}. The reward of each intersection is the same as in the single-intersection setting as in Eq. \ref{eqa:pressure}, and we use the subscript to indicate the reward of the intersection $i$:

\begin{equation}\label{eqa:pressure_marl}
    r_{i} = - P_{i} = - \left| \sum_{(l,m)\in L_{phase}(i)} \left( \frac{n_{i}(l)}{n_{max}(l)} - \frac{n_{i}(m)}{n_{max}(m)} \right) \right|,
\end{equation}

\noindent where 
\begin{itemize}
    \item[--] $P_i$ is the pressure of intersection $i$; 
    \item[--] $L_{phase}(i)$ is a set of combinations of incoming lane $l \in L_{in}(i)$ and outgoing lane $m\in L_{out}(i)$ of each traffic movement controlled by traffic signal phases at intersection $i$;
    \item[--] $n_{i}(l)$ is the number of vehicles in lane $l$;
    \item[--] $n_{max}(l)$ is the capacity of lane $l$, i.e. the maximum number of vehicles the lane can store.
\end{itemize}

{Using pressure as the reward is proved to be able to guide agents to maximize the throughput of the system and thus minimize the travel time of all vehicles \citep{wei2019presslight}. Cooperation among agents is achieved by incorporating traffic information not only from incoming lanes but also from outgoing lanes into the reward calculation, which naturally connects the upstream and downstream intersections.} %\tco{(This reason is not clear. You should compare pressure with other types of rewards, like delay and throughput, and explicityly mentions the difference.)}

%\tco{add some explanation on how the pressure as reward would intrigue coordination among agents automatically}

\subsection{Training algorithm: Asymmetric Advantage Actor-Critic Method}
\label{subsec:a2c_method}

%\tcr{algorithm (replace "strategy" by "algorithm" throughout the paper}
{In this subsection, we will propose a new training algorithm for CVLight, the \emph{Asymmetric Advantage Actor-Critic} (Asym-A2C) method, which can be viewed as an asymmetric variant of the Advantage Actor-Critic method  \citep{mnih2016asynchronous}.}

%{In this subsection, we propose a new set of algorithms named as \emph{Asymmetric A2C} for RL-TSC under scenarios \st{without a full CV penetration rate} \tco{with partial observation}.} Here the A2C refers to the Advantage Actor Critic (A2C) algorithm
{The novelty of Asym-A2C is to leverage its actor-critic architecture to feed asymmetric inputs into the critic and the actor: the critic is fed with full state information while the actor's access is restricted to partially observed information from CVs.} %\st{For an intersection under a scenario without a full penetration rate, the global information is the traffic state of all vehicles and the partial observation is that of detectable CVs.}

\begin{figure}[h]
\centering
\includegraphics[scale=0.25]{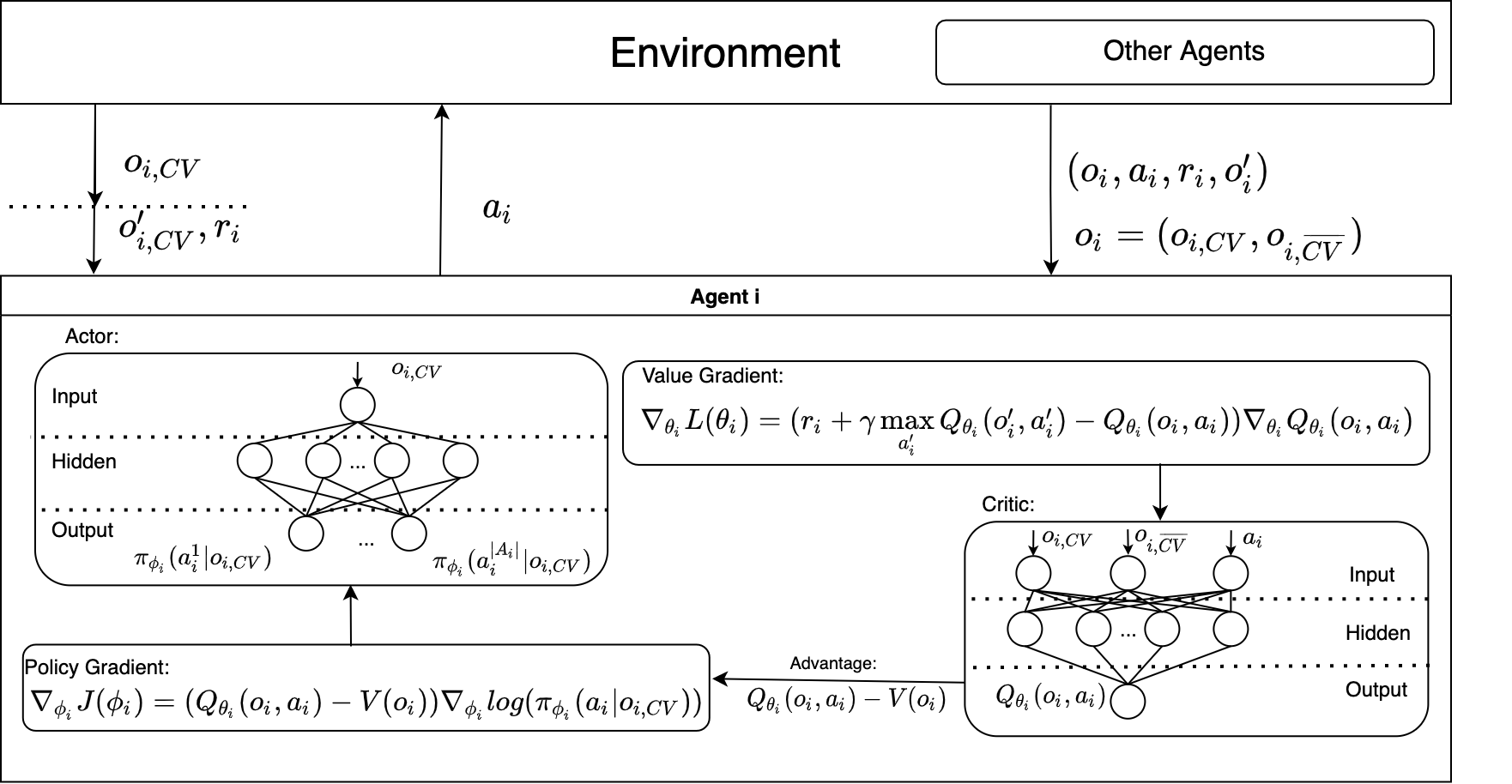}
\caption{An illustration of Asymmetric Advantage Actor-Critic (Asym-A2C) algorithm}

\label{fig:illustration_a2c_local}
\end{figure}
%{Reviewer\#2 1. In Section 2.3.1, the authors claim that each agent is independent in the multi-agent setting. Does it mean that each agent owns its own NN parameters of critic and actor? What is the purpose of doing so? What is the difference if all the agents share the same replay buffer and NN parameters? The problem is, when the controller is implemented for a larger network, then the RL controllers that need to be trained would be numerous rather than one.}

Figure \ref{fig:illustration_a2c_local} details our implementation of Asym-A2C in a multi-intersection scenario where all agents are independent of each other and own the exact same structure as agent $i$ \citep{shou2020marl,shou2020marl_routing}. For the Asym-A2C algorithm, the critic, parametrized by $\theta$, estimates the value function (state value or the state-action value), and the actor subsequently updates its parameters $\phi$ in the direction suggested by the critic. Deep neural networks are used to estimate the value function for the critic as well as the policy model for the actor.
%and {we denote the value network and policy network by $N_\theta$ and $N_\phi$, respectively.} 
{For agent $i$, it observes the full observation $o_{i}$, which is composed of observation from CVs (i.e. $o_{i,CV}$) and that from non-CVs (i.e. $o_{i,\overline{CV}}$). For the actor, we assume it can only receive information from CVs, i.e. the observation for the actor is $o_{i,CV}$. Based on the policy network $\phi_i$, that the actor then chooses one action $a_{i}$ from its action space $A_i$ (we denote the $k$th action in the action space by $a_{i}^{k}$ where $k\in \{1,2,...,|A_i|\}$ and $|A_i|$ is the number of all possible actions). } After the execution of $a_{i}$, the agent can receive a new observation $o_{i}'$ as well as a reward $r_{i}$. An experience tuple $(o_{i},a_{i},r_{i},o_{i}')$ is then stored into the experience replay buffer of agent $i$. In the same way, other agents in the environment store experience tuples into their experience replay buffers independently.
%Experience of each agent is stored in its own experience replay buffer, which stores a set of previous experience tuples in the form of $e = (o,a,r,o')$, where $o$ is the observation, $a$ is the action chosen based on $o$, $r$ is the reward gained, and $o'$ is the next observation as a consequence of action $a$. 

{We assume the training process is off-line and the critic can receive information from both CVs and non-CVs, i.e. $o_{i}$ includes $o_{i, CV}$ and $o_{i, \overline{CV}}$.} During training, experience tuples are sampled from the agent's experience replay buffer and used to update the critic network using the following gradient:

%\tco{wz: \textbf{discussion}. if we move the i to the upper right corner for state etc., should we also move that to the same place for $\theta$ as well as $\phi$? Since intersection is always there while time step is only used in differentiating the experience collected in simulation processes, should we move the t rather than the i to the right upper corner?} \tcr{either way is okay, but be consistent.}

\begin{equation}
\nabla_{\theta_i} L(\theta_i) = ((r_i + \gamma \max_{a'_i} Q_{\theta_i}(o'_i,a'_i))\nabla_{\theta_i} Q_{\theta_i}(o_i,a_i)    
\end{equation}

\noindent where 
\begin{itemize}
    \item[--] $\gamma$ is the discounted factor;
    \item[--] $a'_i$ is the next action that maximizes the value function given the next observation $o'_i$. 
\end{itemize}

Each update on the parameter of the value network $\theta_i$ will allow the agent $i$ to better estimate the value function given the observation and action. 

The policy network is then updated using the gradient computed as follows:
\begin{equation}\label{eqa:policy_gradient}
    \nabla_{\phi_i} J(\phi_i) = \mathbb{E}_{\pi_{\phi_i}}[\nabla_{\phi_i} \log \pi_{\phi_i}(a_i|o_{i,CV})A(o_i,a_i)],
\end{equation}

\noindent where $A(o_i,a_i) = Q_\theta(o_i,a_i)-V(o_i)$, is called the \emph{advantage} of action $a_i$ in the observed state $o_i$. Intuitively, it can be treated as a measure of how much better action $a_i$ is compared to the average. $V(o_i)$ is the baseline, which reduces the gradient size and leads to a much lower variance estimate of the policy gradient~\citep{mnih2016asynchronous}. Each update on the policy network $\phi_i$ will increase the probability of choosing a suitable action so as to achieve a $Q_{\theta}(o_i,a_i)$ higher than the average state value function $V(o_i)$ for the observation $o_i$.

%that enables the agent to achieve

%\subsubsection{Asymmetric A2C}
%\label{subsubsec:intro_a2c_global}

{Now, we summarize key ideas of Asym-A2C as follows:}

\begin{enumerate}
    \item {In the training stage, each critic can evaluate the actor's behavior using information from both CVs and non-CVs at each intersection, while each actor can only observe CV data. Specifically, the input of the critic is composed of traffic observations of CVs and non-CVs. But for the actor, it still can only get information from CVs. To some degree, the critic helps the actor to build a connection between observed CV data and the traffic states of all vehicles;}
    \item {In the execution (or test) stage, actors select their best actions following the learned policies solely using CV data. }
\end{enumerate}

%{Reviewer\#2 2. The descriptions of the A2C variants (Full-sep, Full-total, Full None) in section 2.3.2 are vague. The question here is the separation of the states between CV and non-CV are for the states fed into critic only or for both networks?}

{As a comparison, we also propose the \textit{Symmetric A2C} (denoted by Sym-A2C) where both the actor and the critic receive the same input observation from CV only. We denote the CVLight that utilizes the Asym-A2C and Sym-A2C as \textit{CVLight (Asym)} and \textit{CVLight (Sym)}, respectively. To make a fair comparison, we allow both the CVLight (Asym) agent and the CVLight (Sym) agent to receive rewards based on the traffic states of CVs and non-CVs.} %Note that the Asym-A2C is feasible when historical location data of non-CVs can be extracted from traffic cameras or other sensors for off-line training. } \tco{(I think it is better not to mention it.)}

{As to the settings of our neural networks, for each CVLight agent, the policy neural network 
%$\phi$ \tco{($\phi$ is parameter, not network. To solve the conflict, you should define something like N_$\phi$ in Figure 4 as network first, and use it here.)} 
consists of two hidden layers. Each layer consists of 2$|o_{CV}|$ fully connected neurons and exponential linear unit (ELU) activation functions, where $|o_{CV}|$ is the length of the partial observation vector $o_{CV}$. The value neural network is composed of two hidden layers. Each layer consists of 2$|o|$ fully connected neurons and ELU activation functions. An Adam optimizer \citep{kingma2014adam} with a learning rate 1e-4 is adopted for both value and policy networks. The experience replay buffer size, batch size, and discount rate $\gamma$ are 10,000, 128, and 0.99, respectively.}

%\tco{wz: I move the details of training process here. Or perhaps we can delete this paragraph directly as it is not necessary to be disclosed.}

%{In the training stage, unlimited rounds of simulation will be generated until when the number of updates on neural networks reaches the given threshold (the number of updates is set as 12,000). For the execution stage, the trained RL-TSC model (or other non-learning based TSC algorithms) will be tested for multiple rounds (96 rounds by default) so as to reduce variances in performance measurement. }

\section{Numerical Experiments} %CVLight Model Validation: from Single- to Multi-Intersection
\label{sec:experiments}

%\tcr{This and the next sections need to be restructured completely. We need to clearly highlight the proposed model, which should be prs-duration, and directly frame it on a corridor with multiple intersections and independent learner. All other models are baselines.} \tcr{need to include training, test, validation details as suggested by the reviewers.}

{In this section, we conduct experiments under synthetic road networks and compare the proposed CVLight model with other benchmark models. Experiment settings are introduced in Section \ref{subsec:exp_setting}. In Section \ref{subsec:performance_comp}, we investigate the performance of CVLight and benchmarks under different traffic demand levels and penetration rates. Based on this, we visualize and interpret the learned policies of CVLight in Section \ref{subsec:policy_interpretation}. Furthermore, the sensitivity analysis on the maximum green time (denoted by gmax) and the minimum green time (denoted by gmin) is conducted in Section \ref{subsec:sensitivity_anlysis}. We then discuss the scalability in road network sizes in Section \ref{subsec:scalability}.}

{Hyperparameters in different experiments are summarized in Table~\ref{tab:exp_setting}. The first and second columns list different experiments and corresponding road networks. The remaining columns (3 to 6 for train and 7 to 10 for test) are values of different hyperparameters. All unmentioned hyperparameters are kept as default, such as the 10\% left turn and 10\% right turn proportion. We use bold values to specify the difference of hyperparameters in train and test procedures. In each subsection, we will come back to this table and explain the experiment setting in detail. }

%\begin{landscape}
\begin{table}[H] \centering
\caption {\label{tab:exp_setting} Settings of numerical experiments } 
%\begin{tabular}{p{0.12\linewidth}p{0.1\linewidth}p{0.1\linewidth}p{0.1\linewidth}p{0.1\linewidth}p{0.1\linewidth}p{0.1\linewidth}}
\resizebox{\textwidth}{!}{%
\begin{tabular}{ccccccccccc}
\toprule
\multirow{2}{*}{\textbf{Experiments}} & \multirow{2}{*}{\textbf{Road Networks}} & \multicolumn{4}{c}{\textbf{Train}} & \multicolumn{4}{c}{\textbf{Test}}\\
\cmidrule(lr){3-11}
     &   & Demand$^1$ & gmin (s)  & gmax (s)  & pen$^2$ (\%) & Demand & gmin (s) & gmax (s) & pen (\%)  \\
\midrule
 {\makecell{Generalizability in\\ traffic demands}} & 2-2   &  \textbf{Dynamic} & 7 & 40 & \makecell{\{10,20,30,\\50,70,100 \}} & \makecell{\textbf{Dynamic,}\\\textbf{600,800}} & 7 & 40 & \makecell{\{10,20,30,\\50,70,100 \}} \\
 \makecell{Generalizability in\\ penetration rates} & 2-2 & 600 & 7 & 40 & $\boldsymbol{10}$ & 600 &7 & 40 & \makecell{\{$\boldsymbol{10,20,30,}$ \\ $\boldsymbol{50,100}$\}} \\
% \makecell{Green wave} & 2-2  &  600 & 7 & 40 &  10 & 600 & 7 & 40 &  10  \\ 
\hline
 SA$^3$ on gmin & 2-2   & 600 & \{5,7,10\} & 40 &  \{10,30\} & 600 & \{5,7,10\} & 40 & \{10,30\} \\
  SA on gmax & 2-2 & 600 & 7 & \{40,60,120\} & \{10,30\} & 600 & 7 &  \{40,60,120\} & \{10,30\} \\
 %\hline
 %\makecell{Scalability in \\ road network sizes}  & 5-5  & Dynamic 2 & 7 & 40 & \{10,30\} & Dynamic 2 & 7 & 40 & \{10,30\} \\
 \bottomrule
\end{tabular}}
\begin{tablenotes}
      \item 1. Numbers here represent fixed demands in the unit of vehicles/hour/lane. ``Dynamic'' means the demand is changing according to a fixed pattern, which is detailed in Table \ref{tab:traffic-demand} and Figure \ref{fig:dynamic_traffic_flow_design}.
      \item 2. ``pen'' here represents the CV penetration rate
      \item 3. ``SA'' here represents sensitivity analysis
\end{tablenotes}
\end{table}

\subsection{Experiment Settings}
\label{subsec:exp_setting}

{In this subsection, we introduce the setting of our numerical experiments, including our environment set-up, the measurement of performance, and benchmark algorithms.}

\subsubsection{Environment Set-up}

\bigskip
\noindent \textbf{Road Network}

\begin{figure}[h]
  \subfloat[The layout of an intersection]{
	\begin{minipage}[c][0.9\width]{
	   0.32\textwidth}
	   \centering
	   \includegraphics[width=0.95\textwidth]{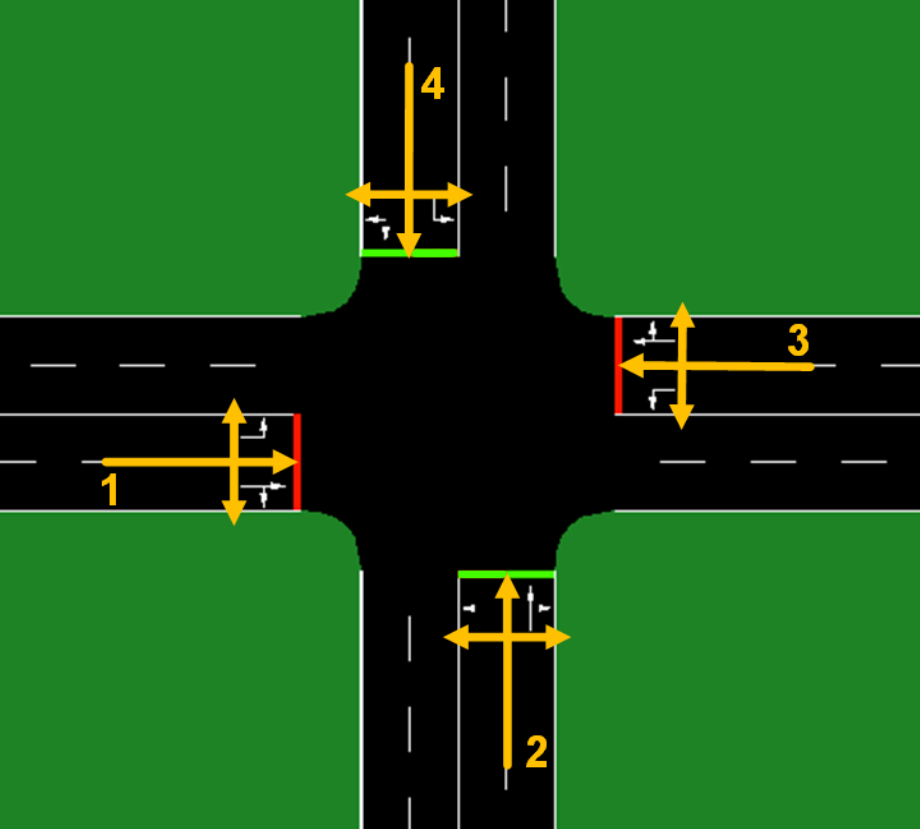}
	\end{minipage}}
 \hfill 	
  \subfloat[2-2 road network layout]{
	\begin{minipage}[c][0.9\width]{
	   0.32\textwidth}
	   \centering
	   \includegraphics[width=0.95\textwidth]{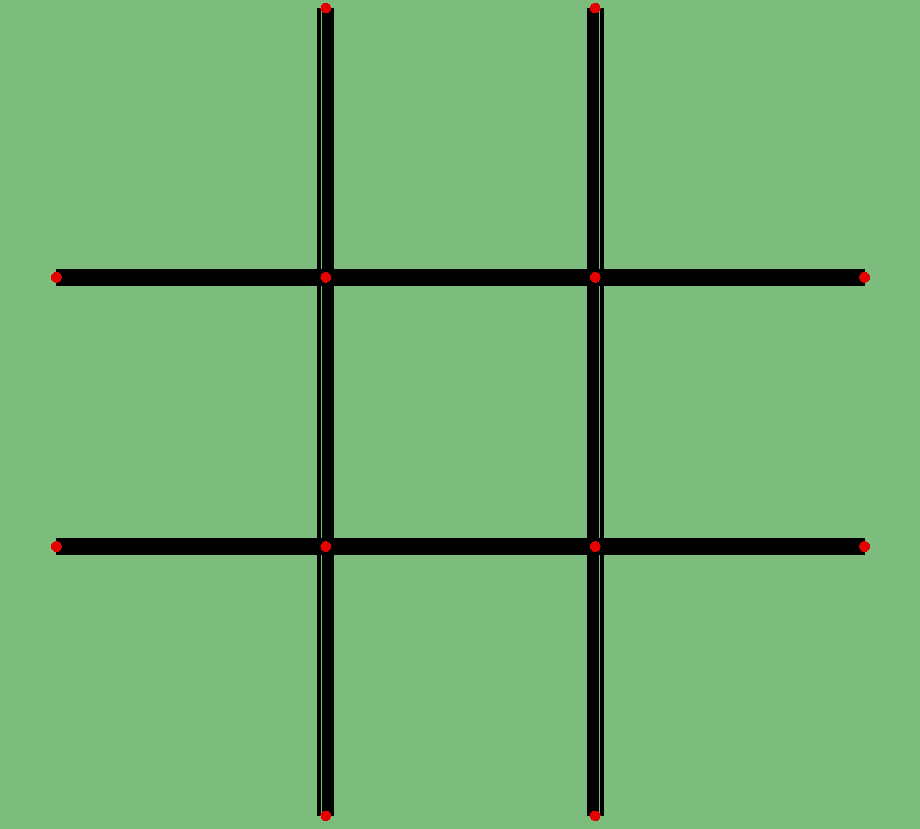}
	\end{minipage}}
 \hfill
  \subfloat[5-5 road network layout]{
	\begin{minipage}[c][0.9\width]{
	   0.32\textwidth}
	   \centering
	   \includegraphics[width=0.95\textwidth]{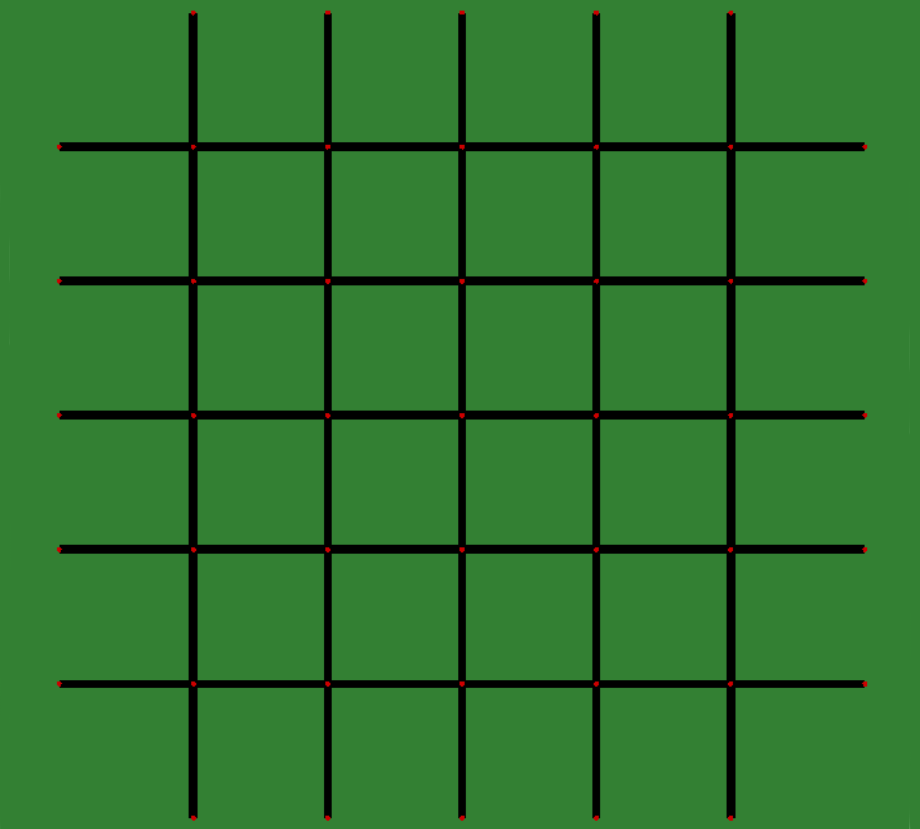}
	\end{minipage}}
\caption{Illustration of Road Network}
\label{fig:road_network}
\end{figure}

{We first introduce a synthetic isolated intersection, based on which we can introduce larger road networks. Figure \ref{fig:road_network}(a) illustrates the structure of a 4-way intersection with two incoming and two outgoing lanes on each way. For each way, each incoming lane consists of one go-straight and right-turn lane and one left-turn lane. The length of each lane is 200 meters and CVs can be detected by the RL-TSC agent when they are within 200 meters of the intersection. }

A 2-phase signal timing plan is used: one phase is the green phase for traffic movement 1 and 3 (represented by arrows in Figure \ref{fig:road_network}(a)); the other is the green phase for traffic movement 2 and 4. Note that traffic movement here is defined as the traffic that is allowed to pass the intersection under a certain phase and travels in the same direction. % \tco{(try a simpler expression. hard to understand)}

%{Reviewer\#1(minor) Page 13, "Vehicles' arrival times follows a binomial distribution." Do you mean the vehicle arrival follows a Poisson process?}
{Two road networks, 2-2 and 5-5, are configured. Each intersection in the road networks follows the same 2-phase signal timing plan as the isolated intersection shown in Figure \ref{fig:road_network}(a). Figure \ref{fig:road_network}(a) illustrates the 2-2 network of 4 intersections and Figure \ref{fig:road_network} (c) illustrates the 5-5 network of 25 intersections. We set the East-West (EW) roads as the arterials and the North-South (NS) roads as the side-streets. The length of each lane in the road network is 200 meters. }

\bigskip
\noindent \textbf{Traffic Demand}

\begin{figure}[H]
  \subfloat[Dynamic traffic demands (train)]{
	\begin{minipage}[c][0.7\width]{
	   0.5\textwidth}
	   \centering
	   \includegraphics[width=1\textwidth]{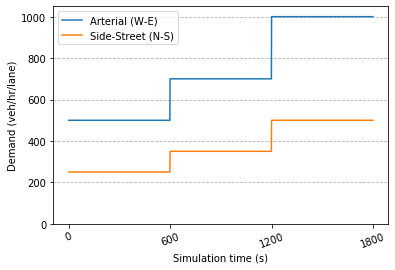}
	\end{minipage}}
 \hfill 	
  \subfloat[Dynamic traffic demands (test)]{
	\begin{minipage}[c][0.7\width]{
	   0.5\textwidth}
	   \centering
	   \includegraphics[width=1.0\textwidth]{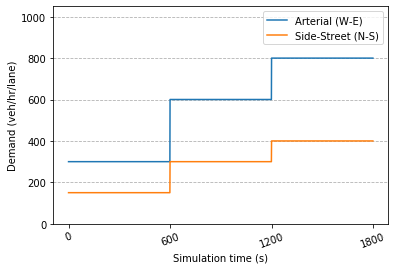}
	\end{minipage}}

\caption{Dynamic traffic demand patterns}
\label{fig:dynamic_traffic_flow_design}
\end{figure}

{All our experiments use synthetic traffic demand data, as shown in Table \ref{tab:traffic-demand} and Figure \ref{fig:dynamic_traffic_flow_design}. For Table \ref{tab:traffic-demand}, the first column lists the traffic demand patterns. The second and third columns detail the traffic demand on the arterial and side-street. Figure \ref{fig:dynamic_traffic_flow_design} presents the dynamic traffic demand patterns, which correspond to the first and second rows in Table \ref{tab:traffic-demand}. The y-axis represents demands and x-axis represents the simulation times. Blue line and orange line mean the demands on the arterials and side-streets, respectively.}

{For each stage in the dynamic traffic demand, the simulation length is 600 seconds. For a fixed demand, the simulation length is 1800 seconds. Vehicles' inter-arrival times follow a binomial distribution by default, i.e. the vehicle arrival process follows a Poisson process. For experiments in this section, both left-turn and right-turn proportions are set to be 10\%.} Given a penetration rate, we randomly assign vehicles to be CVs based on an equal probability of selection. Vehicles that are not chosen to be CVs are the non-CVs in the simulation. 

\begin{table}[H]
\centering
\caption{\label{tab:traffic-demand} Traffic demands}
\begin{tabular}{@{}lccc@{}}
\toprule
\multirow{2}{*}{Demand Patterns} & \multicolumn{2}{l}{Arrival Rate (vehicles/h/lane)}  \\ \cmidrule(lr){2-3}
 & Arterial (EW) & Side-street (NS) & \multicolumn{1}{l}{} \\ \midrule
 Dynamic (train) & 500-700-1000$^1$ & 250-350-500 \\ 
 Dynamic (test) & 300-600-800 & 150-300-400 \\ \midrule
\multirow{2}{*}{Fixed}  & 600 & 300 \\
 & 800 & 400 \\\bottomrule
\end{tabular}
\begin{tablenotes}
      \item {1. The series of numbers represents the demands in different stages in the dynamic traffic demand, which is illustrated in Figure 6. For example, 500-700-1000 means the traffic demand starts from 500 veh/h/lane, increases to 700 veh/h/lane, and ends with 1000 veh/h/lane.}
\end{tablenotes}
\end{table}

%{Reviewer\#1 (minor) (fixed already) Page 13, "each round is one simulation realizat of a given length of time". realizat -> realization?}

\noindent \textbf{SUMO Simulation}

{We conduct experiments in SUMO, an open-source, microscopic, and time-discrete and space-continuous traffic simulation package designed to handle large networks~\citep{behrisch2011sumo,krajzewicz2012recent,SUMO2018}. The simulation of our experiments is round-based, and the duration of each round is 1800 seconds by default in this paper.}

\subsubsection{Performance Measurement}

To measure the performance of a given model in one round of simulation in the execution stage, {we use the average travel delay per vehicle per intersection in a road network as the metric}, denoted by $M_{delay}$, formulated as:
\begin{align}
    M_{delay} &= \frac{1}{|I|}\sum\limits_{i\in I} \Bar{D_i}, \label{equ:metric_delay_1}\\ 
    \Bar{D_i} &= \frac{1}{|T|} \sum\limits_{t=0}^{|T|}(\frac{D_{i,t}}{{\sum\limits_{l\in L_{in}(i)}n_i(l,t)}}), \label{equ:metric_delay_2}\\
    D_{i,t} &= \sum_{l \in L_{in}(i)}  \sum_{v \in V(l, t) } ( t - t_0(v, l) - t_{min}(l)) \label{equ:metric_delay_3},
\end{align}

\noindent where 
\begin{itemize}
    \item[--] $|I|$ is the number of intersections on the road network and $I$ is a set of all intersections in the road network;
    \item[--] $\Bar{D_i}$ is the average travel delay per vehicle at the intersection $i$; 
    \item[--] $|T|$ is the running time of one round in the simulation;
    \item[--] $L_{in}(i)$ is the set of incoming lanes at intersection $i$;
    \item[--] $n_i(l,t)$ is the number of vehicles in lane $l$ at time $t$;
    \item[--] $D_{i,t}$ is cumulative travel delay of all vehicles in all incoming lanes at intersection $i$ at time $t$;
    \item[--] $V(l,t)$ is the set of all vehicles in lane $l$ at time $t$;
    \item[--] $t$ is the time step;
    \item[--]  $t_0(v, l)$ is the time step when vehicle $v$ arrives at lane $l$;
    \item[--] $t_{min}(l)$ is the expected minimum travel time of lane $l$.
\end{itemize}

{The final measurement of performance is the {mean and standard deviation} of $M_{delay}$ of all test rounds. We conduct 96 rounds of tests by default for all experiments.}

%{wz: should we add the description about our neural network structures here?}\tcr{it would help.}

\subsubsection{Benchmarks}
\label{subsec:benchmarks}

%{Reviewer\#1 6. It worth comparing the A2C algorithm to a benchmark algorithm that separates state estimation from signal control to show the advantage of combining the two using the A2C framework. }

To demonstrate the performance of CVLight, we compare it with various popular benchmark TSC algorithms, including both RL based and non-learning based TSC algorithms.

%{Reviewer\#1 7. Page 18, "It worth noting that the experiments of all benchmark algorithms are conducted under 100\% penetration rate, represented as dashed lines in Figure 7." I think most of these benchmark algorithms can also use CV information (potentially with some estimation algorithms). (wz: the original sentence has been re-written as the 'All benchmark algorithms are trained (for RL-TSC) and tested at a 100\% penetration rate (i.e. with full information), which is the scenario they are designed for.')}

%{Reviewer\#2 (minor) (fixed already) b. Page 18, 'It worth noting that…'}

\begin{itemize}
    \item \textbf{Max-pressure.} Max-pressure ~\citep{varaiya2013max} algorithm is a state-of-the-art network-level traffic signal control method, which greedily chooses the phase with the maximum pressure.
    \item \textbf{Adaptive Webster.} Webster's method \citep{webster1958traffic} is one of the classic TSC methods and is widely applied in the real world. To cope with dynamic traffic demands, an adaptive version is developed~\citep{genders2019open}, where the algorithm would periodically update its parameters using the most recent traffic statistics. 
    %in most recent period of time.
     %\item \textbf{Actuated TSC.} Actuated traffic control is another type of classic non-learning TSC method. We apply the SUMO built-in gap-based actuated traffic control as our benchmark.
    % \item \textbf{DDPG} Deep Deterministic Policy Gradient (DDPG) ~\citep{lillicrap2015continuous} algorithm is another classic reinforcement learning algorithm, based on policy gradient method. So the action space of DDPG is different from other benchmarks. It chooses the duration of the pre-defined next phase.
    \item \textbf{Deep Q-Network (DQN) based TSC.}
     DQN ~\citep{mnih2015human} is a classic reinforcement learning algorithm, using a deep neural network to estimate the Q-value function. We adopt this algorithm from \cite{genders2019open}.
     \item \textbf{PressLight.} PressLight ~\citep{wei2019presslight} is another representative state-of-the-art network-level traffic signal control method, based on the Max-pressure.
     
\end{itemize}

{It is worth noting that these benchmark algorithms are not designed for scenarios without the full knowledge of the traffic. To make a fair comparison, we modify the reward calculation for RL based benchmark algorithms: for both PressLight and DQN-based TSC, the reward is calculated based on all vehicles' information.}

\subsection{Performance Comparison}
\label{subsec:performance_comp}

{In this subsection, we compare the performance of CVLigt algorithms and that of baselines under different traffic demands and penetration rates. Specifically, we investigate the generalizability of CVLight in demand patterns and penetration rates: agents are trained under one setting and tested under other settings that they have not seen in training, which are detailed in the first and second rows in Table \ref{tab:exp_setting}.}

\subsubsection{Generalizability in Traffic Demands}

\begin{figure}[H]
  \subfloat[2-2 Dynamic]{
	\begin{minipage}[c][0.4\width]{
	   0.9\textwidth}
	   \centering
	   \includegraphics[width=0.95\textwidth]{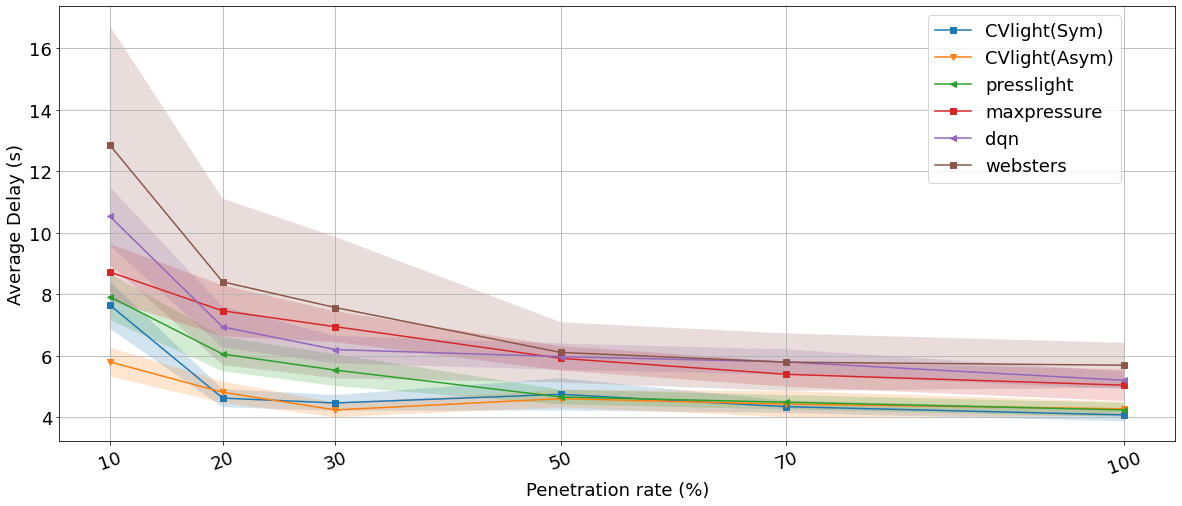}
	\end{minipage}}
\hfill 	
\newline
  \subfloat[2-2 600]{
	\begin{minipage}[c][0.4\width]{
	   0.9\textwidth}
	   \centering
	   \includegraphics[width=0.95\textwidth]{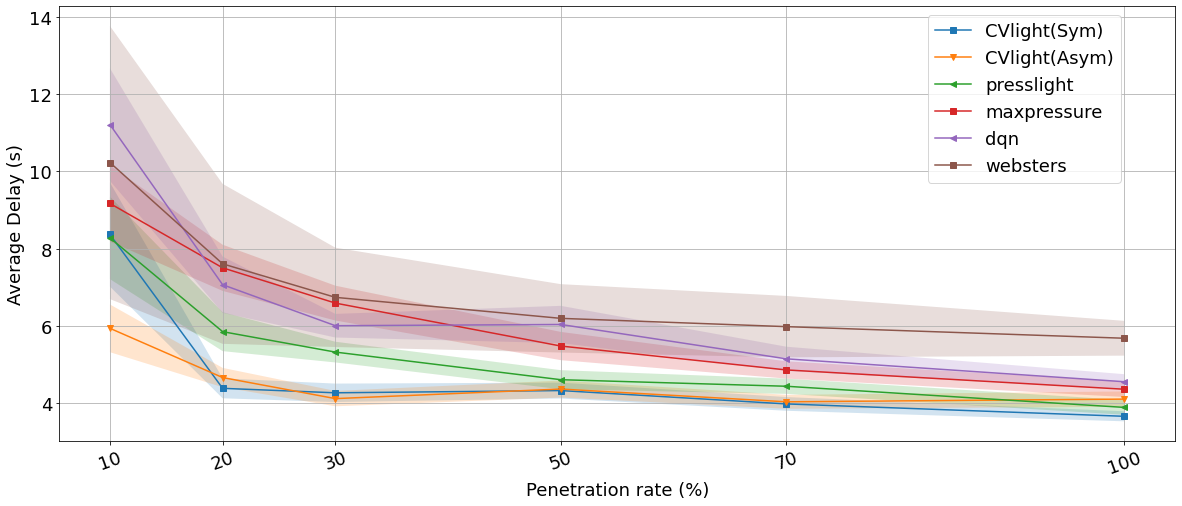}
	\end{minipage}}
\hfill 	
\newline
  \subfloat[2-2 800]{
	\begin{minipage}[c][0.4\width]{
	   0.9\textwidth}
	   \centering
	   \includegraphics[width=0.95\textwidth]{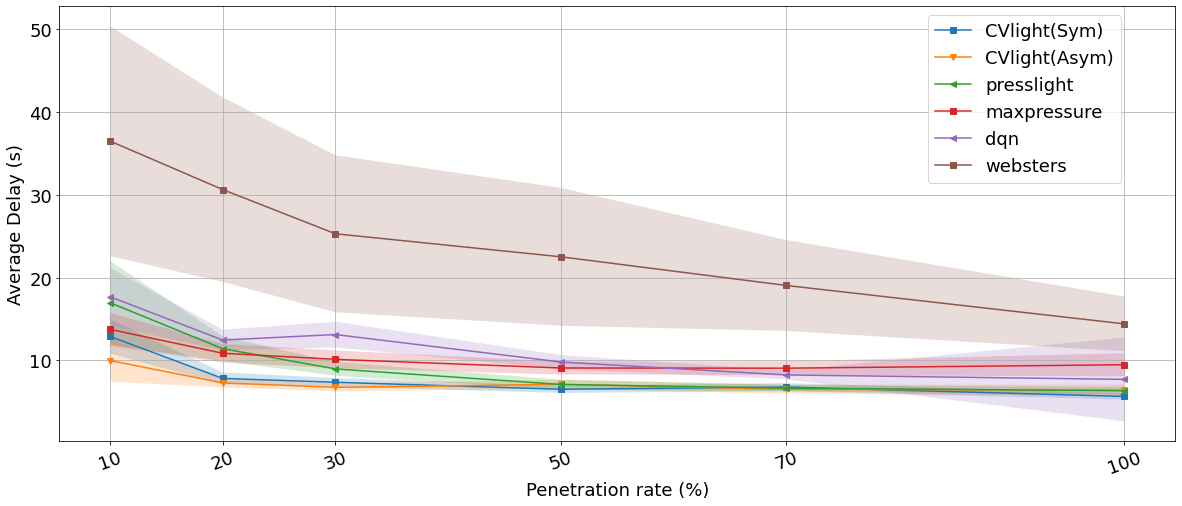}
	\end{minipage}}
\caption{Performance comparison under a 2-by-2 road network}\label{fig:2-2-2_comparison}
\end{figure}

{ To investigate model generalizability in traffic demands, we train our models under one dynamic traffic demand pattern and test them under another dynamic and two fixed traffic demand patterns, as shown in the first row of Table~\ref{tab:exp_setting}. Figure \ref{fig:2-2-2_comparison} compares the performance between CVLight and benchmark algorithms under the 2-2 road network with various penetration rates and traffic demands. In Figure \ref{fig:2-2-2_comparison}, each sub-figure illustrates the performance of CVLight and benchmark algorithms under a certain test demand. The x-axis is the CV penetration rate and the y-axis represents the average delay per vehicle per intersection. The blue, orange, green, red, purple, and brown lines represent the performance of CVLight (Sym), CVLight (Asym), PressLight, Maxpressure, DQN, and Webster's, respectively. The traffic demand is detailed in Table \ref{tab:traffic-demand}. During tests, all models do not have access to the non-CV information.}

{From Figure \ref{fig:2-2-2_comparison}, we can interpret the results from the following two perspectives: }

%(1) Across the three testing scenarios, CVLight (Asym) can outperform all other benchmark algorithms, especially at the penetration rate 10\%, which justifies the advantage of CVLight (Asym). The Asym-A2C enables the CVLight agent to better build connection between the observation from CVs only and the true state with information from all vehicles even when the penetration rate is relatively low. 

%One explanation for the better performance of CVLight (Asym) is that the Asym-A2C enables the CVLight agent to better build connection between the observation from CVs only and the true state with information from all vehicle. With information from both CV and non-CV as input for the critic, the value network in Asym-A2C can better evaluate the action that the actor chosen. Such advantage 

{\textbf{Comparison between CVLight algorithms and other baselines.} CVLight agents can achieve the lowest average delay with small standard deviation values under the three test demand levels, compared to benchmark algorithms. As the train and test traffic demands are different, the good performance of CVLight agents indicates that they can generalize well to unseen traffic demand levels. Specifically, under relatively low penetration rates (below 50\%), CVLight (Asym) and CVLight (Sym) can achieve the most significant advantage over all other benchmark algorithms across all three test scenarios. One explanation for this is the advantage of adding delay and phase duration information into the state, which contain cumulative information over time and provide a more accurate description of the true traffic state than the number of vehicles does under a low penetration rate.}

{\textbf{Comparison between CVLight (Asym) and CVLight (Sym).} Across the three test scenarios, under an extremely low penetration rate (10\%), CVLight (Asym) shows an advantage over CVLight (Sym). A2C-Asym enables the CVLight agent to better build connections between the observation from CVs only and the state from all vehicles. During training, the critic of CVLight (Asym) can access information from all vehicles and therefore evaluate the action chosen by the actor in a more accurate way.}

\begin{figure}[H]
  \subfloat[Training losses of critic 1]{
	\begin{minipage}[c][0.5\width]{
	   0.5\textwidth}
	   \centering
	   \includegraphics[height=0.18\textheight]{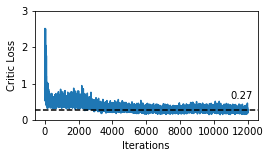}
	\end{minipage}}
 \hfill 	
  \subfloat[Training losses of actor 1]{
	\begin{minipage}[c][0.5\width]{
	   0.5\textwidth}
	   \centering
	   \includegraphics[height=0.18\textheight]{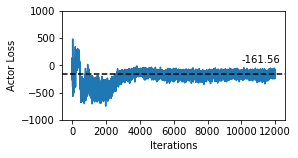}
	\end{minipage}}
 \hfill 	
 \newline
  \subfloat[Training losses of critic 2]{
	\begin{minipage}[c][0.5\width]{
	   0.5\textwidth}
	   \centering
	   \includegraphics[height=0.18\textheight]{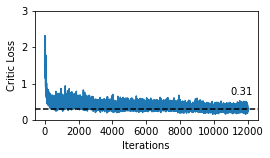}
	\end{minipage}}
 \hfill 	
  \subfloat[Training losses of actor 2]{
	\begin{minipage}[c][0.5\width]{
	   0.5\textwidth}
	   \centering
	   \includegraphics[height=0.18\textheight]{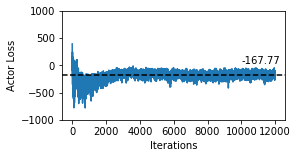}
	\end{minipage}}
 \hfill 	
 \newline
  \subfloat[Training losses of critic 3]{
	\begin{minipage}[c][0.5\width]{
	   0.5\textwidth}
	   \centering
	   \includegraphics[height=0.18\textheight]{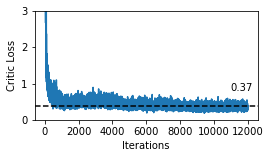}
	\end{minipage}}
 \hfill 	
  \subfloat[Training losses of actor 3]{
	\begin{minipage}[c][0.5\width]{
	   0.5\textwidth}
	   \centering
	   \includegraphics[height=0.18\textheight]{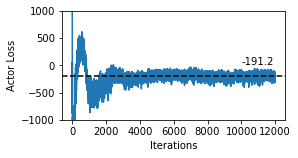}
	\end{minipage}}
 \hfill 	
 \newline
  \subfloat[Training losses of critic 4]{
	\begin{minipage}[c][0.5\width]{
	   0.5\textwidth}
	   \centering
	   \includegraphics[height=0.18\textheight]{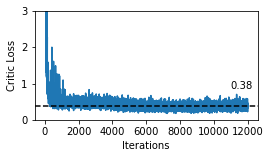}
	\end{minipage}}
 \hfill 	
  \subfloat[Training losses of actor 4]{
	\begin{minipage}[c][0.5\width]{
	   0.5\textwidth}
	   \centering
	   \includegraphics[height=0.18\textheight]{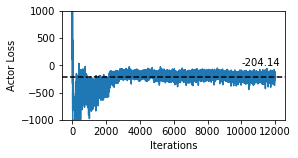}
	\end{minipage}}

\caption{\tcb{Training losses of all CVLight (Asym) agents in the experiment in Figure \ref{fig:2-2-2_comparison} (a)}}
\label{fig:loss}
\end{figure}

\tcb{To better describe training process, Figure \ref{fig:loss} shows the training losses of each actor and critic of all CVLight (Asym) agents in the experiment in Figure \ref{fig:2-2-2_comparison} (a). 
In Figure \ref{fig:loss}, the x-axis is the training iteration index, and the y-axis is the loss value. The blue lines represent the progressions of the training losses of the critics (left column)  and actors (right column), each row representing one agent. The dashed black lines and numbers represent the average loss values in the last 1,000 iterations. From Figure \ref{fig:loss}, all actors and critics converge well.} 

%(2) By comparing Figure \ref{fig:2-2-2_comparison} (a) and (b), when the traffic demand become heavier, the difference in performance of a certain algorithm under a high penetration rate (e.g. 100\%) and performance under a low penetration rate (e.g. 10\%) gets smaller. Under a 

%{\textbf{Generalizability of CVLight in traffic demand.} CVLight agents generalize well to traffic demand levels they have not seen in training: compared to benchmark algorithms, they can achieve the lowest average delay with small standard deviation values under various test demand levels.} \tco{(not a section that should be in parallel to others. in a non-suitable place.)}

%(3) [Influence of the demand level on performance] Under a relatively lighter traffic demand, as shown in Figure \ref{fig:2-2-2_comparison} (b), the increase of the delay as penetration decreasing

%\tco{During testing, all models do not have access to the non-CV information. Figure \ref{xxx} shows results of a 2-2 road network for our proposed method and baselines. The blue, orange, green, red and purple lines represent ... . The specific experiment details can be found in Section xxx. As shown in Figure \ref{xxx}, xxx performs the best, which justifies the advantage of CVLight. When the penetration rate is 10\%, ... . } \tco{Figure \ref{xxx} shows that ... .}\tco{ One explanation for the better performance of CVLight is ..., especially when the penetration rate is extremely low, e.g., 10\%. During training, with traffic information from both CVs and non-CVs, the critic of CVLight can ... .}

\subsubsection{Generalizability in Penetration Rates} 

{We evaluate the generalizability of our proposed model under unseen penetration rates: we train our models under 10\% penetration rate and test them under 10\%, 30\%, 50\%, and 100\% penetration rates, as shown in the second row of Table~\ref{tab:exp_setting}. The results are shown in Table~\ref{tab:pen-generalization}. The first column lists the names of algorithms and the second to the sixth columns present the average delay achieved by each algorithm under 10\%, 20\%, 30\%, 50\%, and 100\% penetration rates, respectively. Each entry in the table is the mean value and standard deviation of average delays across all test rounds. We can interpret the results as below:} 

{CVLight (Asym) and CVLight (Sym) show good generalizability in penetration rates: their performance under an unseen high penetration rate, e.g. 100\%, can achieve the same level as their performance under 10\%, while PressLight and DQN agents fail to generalize well to such a high penetration rate. The good performance of CVLight can be contributed to the state design, which includes both current traffic information, e.g. the number of vehicles in each lane, and time-cumulative information, e.g. average delay in each lane. CVLights significantly decrease the average delay compared to baselines across all test penetration rates. Specifically, CVLight (Asym) achieves the minimal average delay in general across all test penetration rates, which can be contributed to the Asym-A2C.} 
%although the difference in performance between CVLight (Asym) and CVLight (Sym) is not as significant as that between CVLights and other benchmarks.} \tco{(dont mention the insignificance. you can mention the advantage of CVlight(asym) can be attributed to the asym A2C)} 

\begin{table}[H]
\renewcommand{\thetable}{\arabic{table}}
\centering
\caption{\label{tab:pen-generalization} Performance comparison on generalizability in penetration rate}
%\resizebox{\columnwidth}{!}{%
\begin{tabular}{cccccccccccc}
\toprule
\multicolumn{1}{c}{\multirow{2}{*}{\begin{tabular}[c]{@{}c@{}} Algorithms \end{tabular}}} & \multicolumn{5}{c}{CV penetration rate} \\ 
\cline{2-7}
\multicolumn{1}{c}{}  & 10\% & 20\% & 30\% & 50\% & 100\%  \\ 
\midrule
\makecell{CVLight \\ (Asym)}  & $\textbf{4.53}\pm0.36$ & $\textbf{4.34}\pm0.30$ & $\textbf{4.24}\pm0.30$ & $\textbf{4.14}\pm0.25$ & $\textbf{4.09}\pm0.27$ \\
\midrule
\makecell{CVLight \\ (Sym)}  & $4.58\pm0.39$ & $4.43\pm0.35$ & $4.33\pm0.29$ & $4.30\pm0.29$ & $4.25\pm0.31$ \\
\midrule
\makecell{Presslight}  & $4.94\pm0.33$ & $5.38\pm0.36$ & $5.53\pm0.47$ & $6.15\pm0.65$ & $9.13\pm1.99$ \\
\midrule
DQN  & $5.39\pm0.36$ & $5.62\pm0.42$ & $5.87\pm0.99$ & $6.98\pm4.75$ & $33.37\pm17.50$ \\
\bottomrule
\end{tabular}
%}
\end{table}

%\tcr{Where is interpretability responding to Reviewer 3? Where is the greenwave part?}

\subsection{Learned Policy Interpretation}
\label{subsec:policy_interpretation}

\begin{figure}[h]
  \subfloat[CVLight (Asym) learned policy]{
	\begin{minipage}[c][0.8\width]{
	   0.5\textwidth}
	   \centering
	   \includegraphics[width=0.95\textwidth]{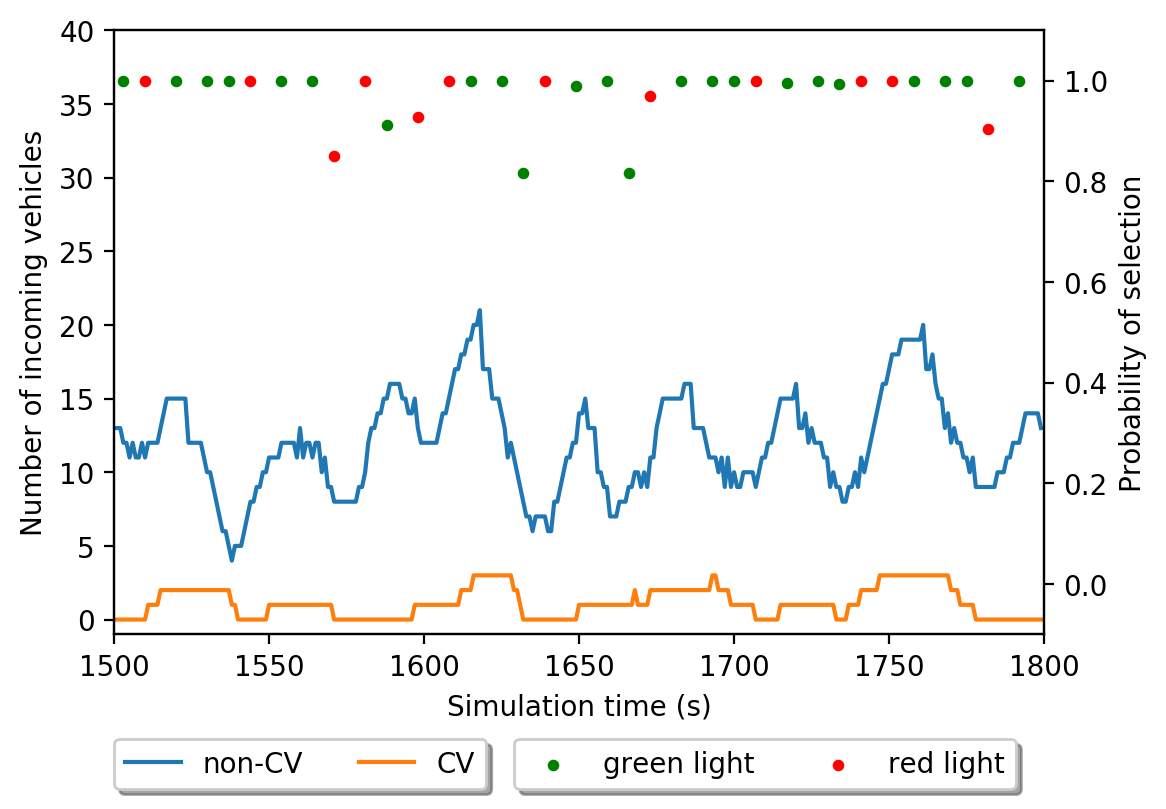}
	\end{minipage}}
\hfill 	
  \subfloat[CVLight (Sym) learned policy]{
	\begin{minipage}[c][0.8\width]{
	   0.5\textwidth}
	   \centering
	   \includegraphics[width=0.95\textwidth]{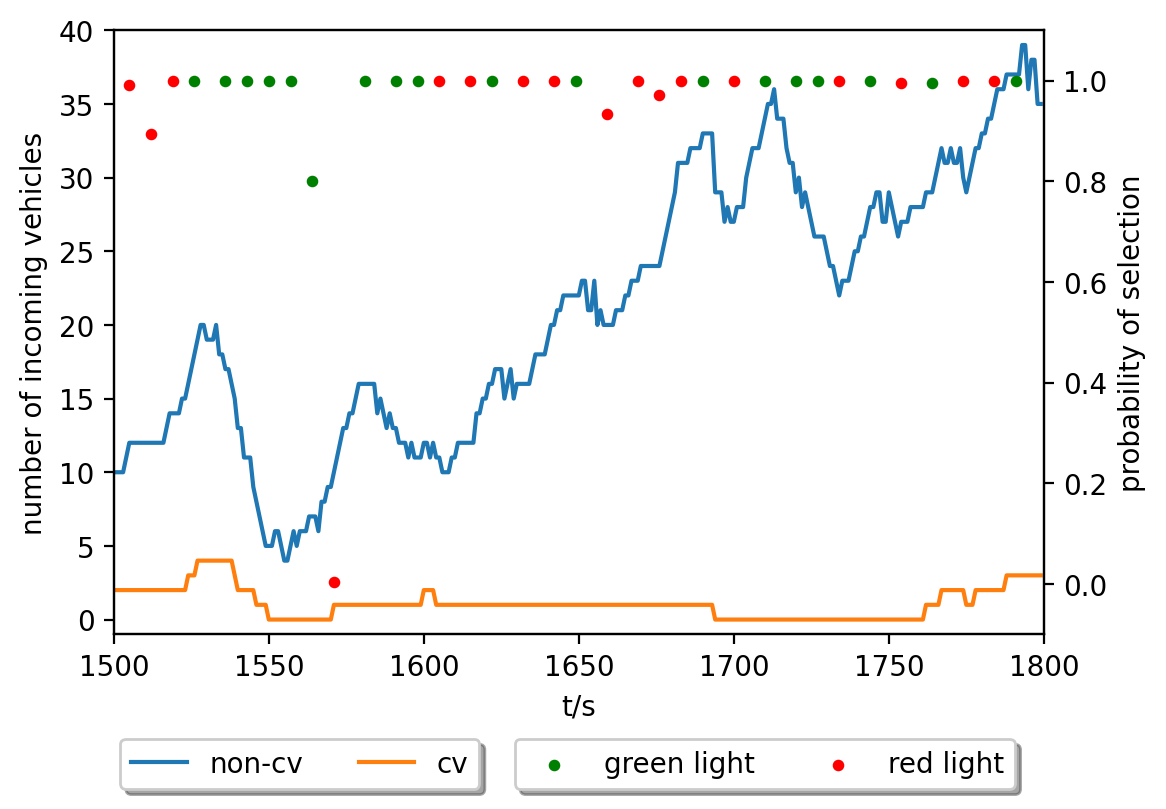}
	\end{minipage}}
\caption{Policy visualization for CVLight (Asym) and CVLight (Sym) under 2-2 with dynamic traffic demand and 10\% penetration rate}
\label{fig:viz_policy}
\end{figure}

{Learned policies of CVLight (Asym) and CVLight (Sym) are illustrated as in Figure~\ref{fig:viz_policy}. Those policies correspond to the 10\% penetration rate case in Figure~\ref{fig:2-2-2_comparison}(a). Figure~\ref{fig:viz_policy} presents the visualization of the learned policy of CVLight agents, where the left and right sub-figures are results of CVLight(Asym) and CVLight(Sym), respectively. The x-axis is the simulation time step, the left y-axis is the number of vehicles that are waiting in the incoming lane of the arterial, and the right y-axis is the probability of choosing a green phase or a red phase (i.e. not selecting the green phase) in the arterial direction. 
%Figure~\ref{fig:viz_policy}(a) and (b) detail the learned policies and the situation in the decision-making process. 
Actors of CVLight agents can only observe CVs, i.e. orange lines in the figure, and make decisions based on that information. The green dot or red dot represents the action (i.e. the next phase to be executed) that the agent chose at every decision time and the position of each dot shows the probability of selecting such action based on the learned policy. The red dot at around 1570 seconds in Figure \ref{fig:viz_policy}(b) is caused by the enforcement of the maximum green time (40 seconds), considering the 6 green dots before the red dot and the 7 seconds interval between each pair of dots.  We can analyze the result as below:} 

{Comparing Figure~\ref{fig:viz_policy}(a) to Figure~\ref{fig:viz_policy}(b), CVLight (Asym) agent can better balance the traffic flow in both the arterial and the side-street directions and avoid congestion: it learns to give longer green phases to the arterial direction while switching to the side-street direction for a short time when it observes few CVs in the arterial direction. Specifically, we can compare the actions of CVLight (Asym) and CVLight (Sym) at simulation time 1550 seconds when both agents do not observe any CVs in the arterial direction: CVLight (Asym) chooses to give green phase to the side-street while CVLight (Sym) continues to give green phase to the arterial until the enforcement of the maximum green time at around 1570 seconds. Such extreme behavior of CVLight (Sym), i.e. keeping a certain phase for a long time, can explain the performance difference between CVLight (Asym) and CVLight (Sym) under the 10\% penetration rate as shown in Figure \ref{fig:2-2-2_comparison}(a). This demonstrates the effectiveness of Asym-A2C: by incorporating non-CV information into training, CVLight (Asym) agent can learn to make better decisions under a low penetration rate scenario.}

\begin{figure}[H]
  \subfloat[10\% penetration rate]{
	\begin{minipage}[c][0.3\width]{
	   0.9\textwidth}
	   \centering
	   \includegraphics[width=\textwidth]{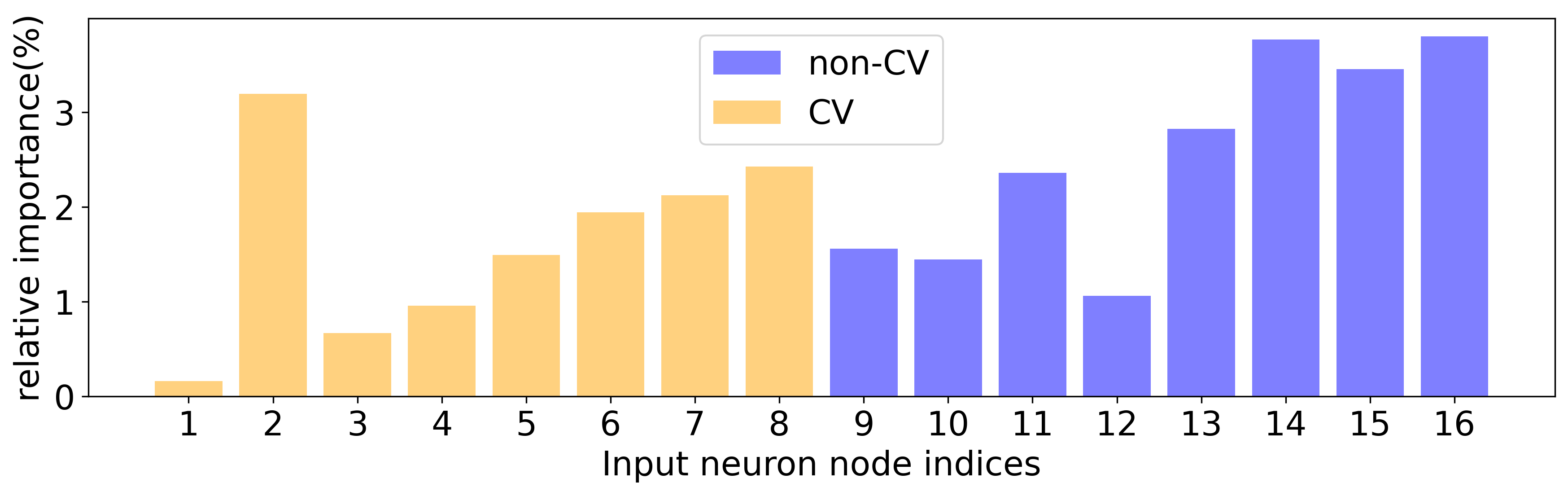}
	\end{minipage}}
\hfill 	
\newline
  \subfloat[100\% penetration rate]{
	\begin{minipage}[c][0.3\width]{
	   0.9\textwidth}
	   \centering
	   \includegraphics[width=\textwidth]{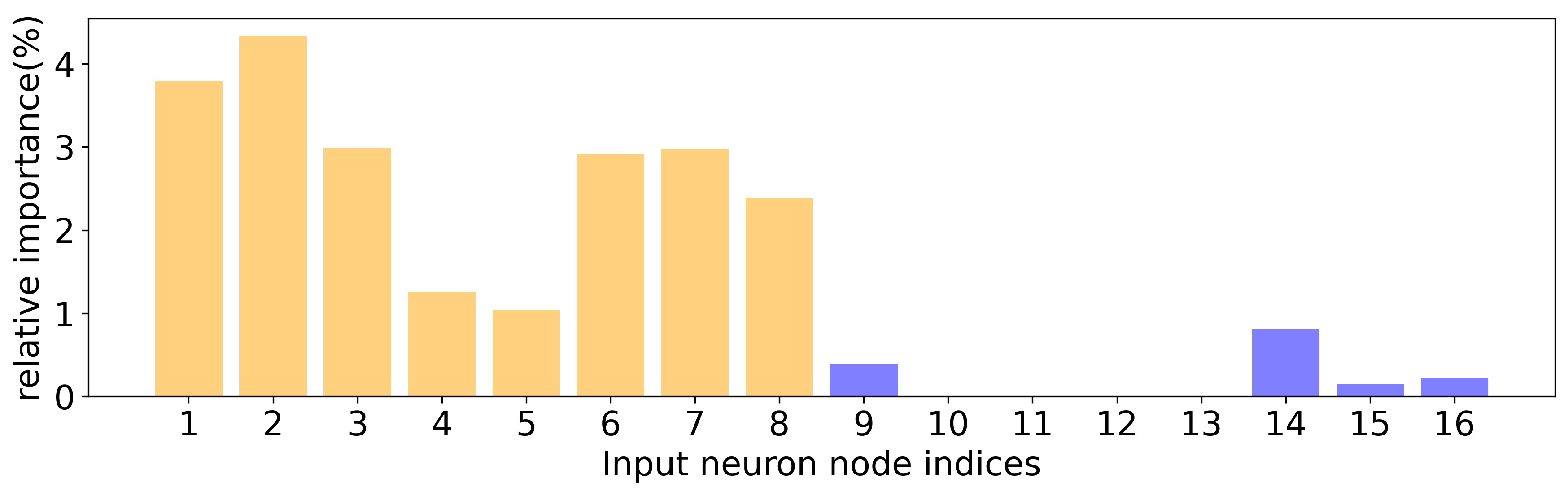}
	\end{minipage}}
\caption{Relative importance of delay related neurons in the input layer of the CVLight (Asym) critic neural network}\label{fig:nnweight}
\end{figure}

 {To further investigate the effectiveness of Asym-A2C, Figure~\ref{fig:nnweight} illustrates the relative importance \citep{gevrey2003review} of each delay related neuron in the input layer of the CVLight (Asym) critic neural network. For each input neuron indexed by $j$, the relative importance, denoted by $RI(\%)_j$, is presented as below: }
 
 \begin{align}
     RI(\%)_j &= \frac{\sum\limits_{h=1}^{N_h}Q_{jh}}{\sum\limits_{h=1}^{N_h}\sum\limits_{j=1}^{N_j}Q_{jh}} \times 100, \\
     Q_{jh}&=\frac{|W_{jh}|}{\sum\limits_{j=1}^{N_j}|W_{jh}|},
 \end{align}
     
     where
     \begin{itemize}
        \item[--] $j$ is the input neuron index;
         \item[--] $h$ is the hidden neuron index;
         \item[--] $N_j$ is the number of neurons in the input layer;
         \item[--] $N_h$ is the number of neurons in the next hidden layer;
         \item[--] $|W_{jh}|$ is the absolute value of the input-hidden layer connection weight between input neuron $j$ and hidden neuron $h$.
     \end{itemize}
 
 {The two sub-figures in Figure~\ref{fig:nnweight} correspond to the experiments in Figure \ref{fig:2-2-2_comparison}(a) under penetration rates 10\% and 100\%, respectively. The x-axis represents the indicies of delay related neurons in the input layer, where the 1st to 8th neurons relate to the CV delay information (each neuron corresponds to one incoming lane of the intersection) and the 9th to 16th neurons relate to the non-CV delay information. The y-axis represents the relative importance of the inputs. The orange bars represent the relative importance of neurons corresponding to the CV delay and the purple bars represent the  relative importance of neurons corresponding to the non-CV delay. To highlight the difference between CV and non-CV delay related neurons, we generate the relative importance using weights with absolute values higher than 0.267, which is the limit of the He uniform weights initialization \citep{he2015delving} in our case (the limit is the $\sqrt{6/|o_{i}|}$, where the $|o_i|$ is the number of neurons in the input layer of the critic neural network of agent $i$, which is 84 in this experiment).}

{From Figure \ref{fig:nnweight}(a), the non-CV delay related neurons contain higher values of the relative importance than the CV delay related neurons do in general, which shows that non-CV delay information gains more attention from the agent under the 10\% penetration rate scenario. Figure \ref{fig:nnweight}(b) shows that the CV related neurons contain higher values in the relative importance than the non-CV related neurons do, which means CV delay information is more representative under the 100\% penetration rate. In summary, relative importance of delay related input layer neurons shows that the critic of CVLight (Asym) agent does learn to rely more on the non-CV delay information under a low penetration rate scenario. This also explains the superiority of CVLight (Asym) over CVLight (Sym) under low penetration rate scenarios.}

%\tco{insert a new figure for Green Wave visualization. 1) introduce the experiment hyperparameters; 2) introduce the figure; 3) interprete the figure.}

\subsection{Sensitivity Analysis}\label{subsec:sensitivity_anlysis}

{In this subsection, we conduct sensitivity analysis (SA) on our proposed model using numerical experiments. We investigate the robustness of our proposed model under different control intervals (minimum green times) and maximum green times.}

 {All experiments conducted in this section are based on the 2-2 road network with a fixed traffic demand, i.e. 600 veh/h/lane for the arterial and 300 veh/h/lane for the side-street. We only focus on low penetration rates of 10\% and 30\% where average delays of different models are the most distinct according to Figure~\ref{fig:2-2-2_comparison}.}

\subsubsection{Control Interval}
{We conduct experiments under 3 different control intervals: 5, 7, and 10 seconds, which are shown as the third row of Table~\ref{tab:exp_setting}. The results are shown in Table~\ref{tab:gmin}. The first column is the names of different models. The remaining columns are organized in two hierarchies. The top level columns are three different values of control intervals. The second level columns are two penetration rates. Each entry is the mean and standard deviation of the average delay after 96 test rounds. For each column, the minimal average delay is in bold. We can interpret the results in three perspectives:}

{\textbf{Influence of control interval on the performance of CVLight.} Comparing the performance of CVLight under gmin 5 and gmin 7, a too short control interval, i.e. gmin 5, deteriorates the performance of CVLight under a low penetration rate, as agents switch the phase more frequently than necessary when CVs are not observed. Due to lack of flexibility, CVLight agents also suffer from a too long control interval when they gain more information from CVs, comparing the performance of CVLight under gmin 7 and gmin 10 with the 30\% penetration rate.}

{\textbf{Comparison between CVLight algorithms and baselines}. CVLight (Asym) and CVLight (Sym) significantly decrease the average delay compared to baselines across all control intervals and penetration rates.}

{\textbf{Comparison between CVLight (Asym) and CVLight (Sym)}. CVLight (Asym) achieves the minimal average delay in general. CVLight (Asym) has a notable improvement compared to CVLight (Sym), in which the biggest improvement is 23.4\% (5 seconds Gmin and 10\% penetration rate).}

\begin{table}[H]
\renewcommand{\thetable}{\arabic{table}}
\centering
\caption{\label{tab:gmin} Performance comparison under different control intervals (gmin)}

%\resizebox{\columnwidth}{!}{%
\begin{tabular}{cccccccccccc}
\toprule
\multicolumn{1}{c}{\multirow{2}{*}{\begin{tabular}[c]{@{}c@{}} Algorithms \end{tabular}}} & \multicolumn{2}{c}{Gmin 5} & \multicolumn{2}{c}{Gmin 7} & \multicolumn{2}{c}{Gmin 10} \\ 
\cline{2-7}
\multicolumn{1}{c}{}  & $10\%^1$ & 30\%  & 10\% & 30\% & 10\% & 30\% \\ 
\midrule
\makecell{CVLight \\ (Asym)}  & $\textbf{8.27}\pm0.98$ & $\textbf{4.48} \pm 0.24$ & $\textbf{4.53}\pm0.36$ & $\textbf{3.82}\pm0.24$ & $\textbf{4.35} \pm 0.20$ & $\textbf{3.92} \pm 0.18$ \\ 
\midrule
\makecell{CVLight \\ (Sym)}  & $10.8 \pm 1.27$ & $5.28\pm0.38$ & $4.58\pm0.39$ & $3.90\pm0.25$ & $4.50\pm0.25$ & $4.23\pm0.18$ \\ 
\midrule
PressLight  & $10.67\pm1.80$ & $6.03\pm0.46$ & $4.94\pm0.33$ & $4.85\pm0.35$ & $4.81\pm0.32$ & $4.75\pm0.27$ \\ 
\midrule
DQN  & $17.22\pm6.88$ & $11.27\pm1.50$ & $5.39\pm0.36$ & $5.04\pm0.31$ & $9.15\pm1.21$ & $5.04\pm0.27$ \\ 
\midrule
Maxpressure & $9.22\pm0.98$ & $7.11\pm0.68$ & $9.07\pm0.95$ & $6.47\pm0.37$ & $6.98\pm2.08$ & $5.45\pm0.48$ \\ 
\bottomrule
\end{tabular}
%}
\begin{tablenotes}
      \small
      \item 1. The percentages here refer to the CV penetration rates.
\end{tablenotes}

\end{table}

\subsubsection{Maximum Green Time}
{We conduct experiments under 3 different maximum green times: 40, 60, and 120 seconds, which are shown in the fourth row of Table~\ref{tab:exp_setting}. The results are shown in Table~\ref{tab:gmax}. It shares the same structure as Table~\ref{tab:gmin} except for the top level of the columns, which are values of the maximum green time. We can interpret the results from three perspectives:}

{\textbf{Influence of the maximum green time on the performance of CVLight.} Under the 10\% penetration rate scenario, a too long maximum green time can influence the performance of CVLight as agents might continue to give a certain direction green phase for too long when CVs in another direction are not observed. 
Comparing the performance of CVLight algorithms under the 30\% penetration rate across the three gmax values and the its performance under the 10\% penetration rate across the three gmax values, we can see that the effect of gmax on the performance will not be that significant when penetration rate gets higher as agents can make better decisions with more information from CVs.}

%The difference in the performance of CVLight algorithms under 30\% with different gmax values is smaller than the difference in the performance of CVLight under 10\% penetration rate with different gmax values \tco{(too complicated to understand. Just delete it and   the following "This shows that")}. This shows that the effect of gmax on the performance will not be that significant when penetration rate gets higher as agents can make better decisions with more information from CVs.}

{\textbf{Comparison between CVLight algorithms and baselines}. CVLight (Asym) and CVLight (Sym) significantly decrease the average delay compared to baselines across all maximum green times and penetration rates.}

{\textbf{Comparison between CVLight (Asym) and CVLight (Sym)}. Across the three max green times, CVLight (Asym) notablely outperforms CVLight (Sym) in terms of the average delay.}

%There are two exceptions: 10\% penetration rate with 40 seconds maximum green time, and 30 \% penetration rate with 60 seconds maximum green time. For those two exceptions, however, the advantages of CVLight (Sym) over CFLight (Asym) are little, which are 1.2\% and 1.4\%, respectively. CVLight (Asym) notably outperforms CVLight (Sym).

\begin{table}[H]
\renewcommand{\thetable}{\arabic{table}}
\centering
\caption{\label{tab:gmax} Performance comparison under different maximum green times (gmax)}
%\resizebox{\columnwidth}{!}{%
\begin{tabular}{cccccccccccc}
\toprule
\multicolumn{1}{c}{\multirow{2}{*}{\begin{tabular}[c]{@{}c@{}} Algorithms \end{tabular}}} & \multicolumn{2}{c}{Gmax 40} & \multicolumn{2}{c}{Gmax 60} & \multicolumn{2}{c}{Gmax 120} \\ 
\cline{2-7}
\multicolumn{1}{c}{}  & $10\%^1$ & 30\%  & 10\% & 30\% & 10\% & 30\% \\ 
\midrule
\makecell{CVLight \\ (Asym)}  & $\textbf{4.53}\pm0.36$ & $\textbf{3.82}\pm0.24$ & $\textbf{4.71}\pm0.38$ & $\textbf{3.92}\pm0.22$ & $\textbf{5.18} \pm 0.55$ & $\textbf{3.88} \pm 0.18$ \\ 
\midrule
\makecell{CVLight \\ (Sym)}  & $4.58\pm0.39$ & $3.90\pm0.25$ & $5.12\pm0.51$ & $4.00\pm0.23$ & $5.53\pm0.95$ & $4.06\pm0.41$ \\ 
\midrule
PressLight  & $4.94\pm0.33$ & $4.85\pm0.35$ & $6.74\pm0.95$ & $5.30\pm0.41$ & $6.49\pm1.07$ & $4.98\pm0.46$ \\ 
\midrule
DQN  &  $5.39\pm0.36$ & $5.04\pm0.31$ & $8.22\pm1.29$ & $6.60\pm0.47$ & $10.77\pm 2.91$ & $6.77\pm 0.75$ \\ 
\midrule
Maxpressure & $9.07\pm0.95$ & $6.47\pm0.37$ & $8.89\pm0.96$ & $6.50\pm0.45$ & $8.88\pm0.97$ & $6.54\pm0.47$ \\ 
\bottomrule
\end{tabular}
%}
\begin{tablenotes}
      \small
      \item 1. The percentages here refer to the CV penetration rates.
\end{tablenotes}

\end{table}

\subsection{Scalability in Road Network Sizes}
\label{subsec:scalability}

\tcb{In this subsection, we demonstrate the scalability of CVLight in terms of road network sizes by applying pre-trained CVLight for a 5-5 road network.} 

%\subsubsection{Scalability in Road Network Sizes}

%\tcr{how about 3-by-3?}

\tcb{To further improve the scalability of CVLight, we use a pre-train technique to speed up the convergence in neural networks training. The pre-train technique allows RL agents to load the pre-trained RL models, as long as they share the same neural network structures, and continue the training until convergence.}  \tcb{ Under the 10\% or the 30\% penetration rate, CVLight agents are pre-trained on the 2-2 road network with a fixed traffic demand pattern for 12,000 iterations, and then are loaded and trained for another 12,000 iterations on the 5-5 road network with a dynamic traffic demand pattern. To better demonstrate the effectiveness of the pre-train technique, we design traffic demand patterns that are significantly different between the pre-train stage and the training stage, as summarized in Table \ref{tab:scale_exp_demand} and Figure \ref{fig:scale_exp_flow} below. For Table \ref{tab:scale_exp_demand}, the first column gives the stages, the second column presents the road network, the third column indicates the traffic demand patterns, and the fourth to sixth columns detail the arrival rate in each direction. Such traffic patterns are also illustrated in Figure \ref{fig:scale_exp_flow} where arrows represent the directions of arriving vehicles.}
%the last row in Table \ref{tab:exp_setting}}.

\begin{table}[H]
\centering
\caption{\label{tab:scale_exp_demand} Traffic demands}
\begin{tabular}{@{}lccccccc@{}}
\toprule
\multirow{3}{*}{Stages} & \multirow{3}{*}{\makecell{Road \\ Network}} & \multirow{3}{*}{\makecell{Demand \\Patterns}} &  \multicolumn{4}{c}{Arrival Rate (vehicles/h/lane)}  \\ \cmidrule(lr){4-7}
 & & & \makecell{Arterial \\ (E-W)} & \makecell{Arterial \\ (W-E)} & \makecell{Side-street \\ (N-S)} & \makecell{Side-street \\ (S-N)} \\ \midrule
Pre-train & 2-2& Fixed  & 300 & 300 & 150 & 150 \\\midrule
Training & 5-5& Dynamic & 500-700-1000$^1$ & 250-350-500 & 250-350-500 & 125-175-250 \\ \bottomrule
\end{tabular}
\begin{tablenotes}
      \item {1. The series of numbers represents the demands in different stages in the dynamic traffic demand. For example, 500-700-1000 means the traffic demand starts from 500 veh/h/lane, increases to 700 veh/h/lane, and ends with 1000 veh/h/lane.}
\end{tablenotes}
\end{table}

\begin{figure}[h]
  \subfloat[2-2 road network for pre-train]{
	\begin{minipage}[c][0.9\width]{
	   0.5\textwidth}
	   \centering
	   \includegraphics[width=0.95\textwidth]{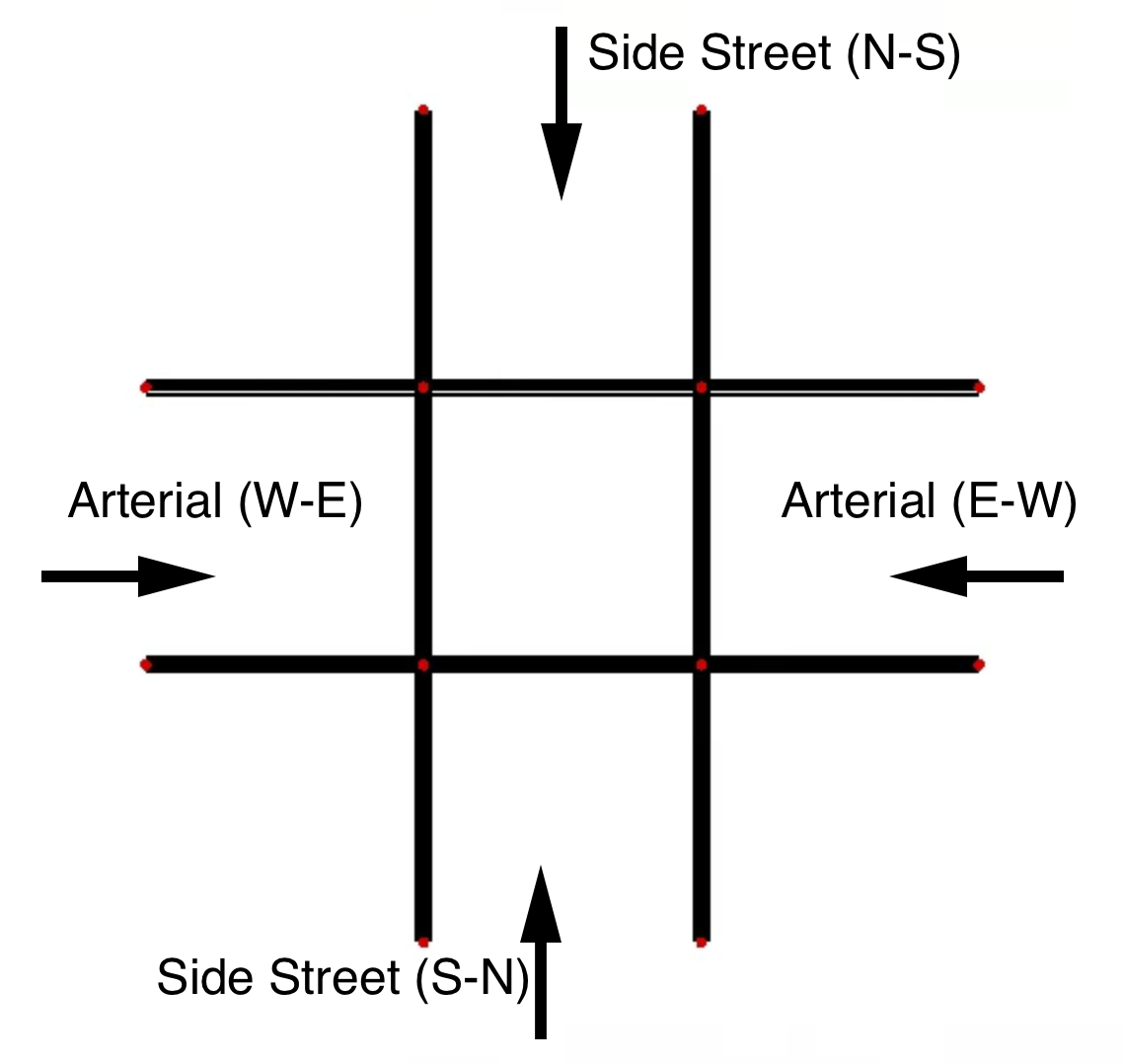}
	\end{minipage}}
\hfill 	
  \subfloat[5-5 road network for training]{
	\begin{minipage}[c][0.9\width]{
	   0.5\textwidth}
	   \centering
	   \includegraphics[width=0.95\textwidth]{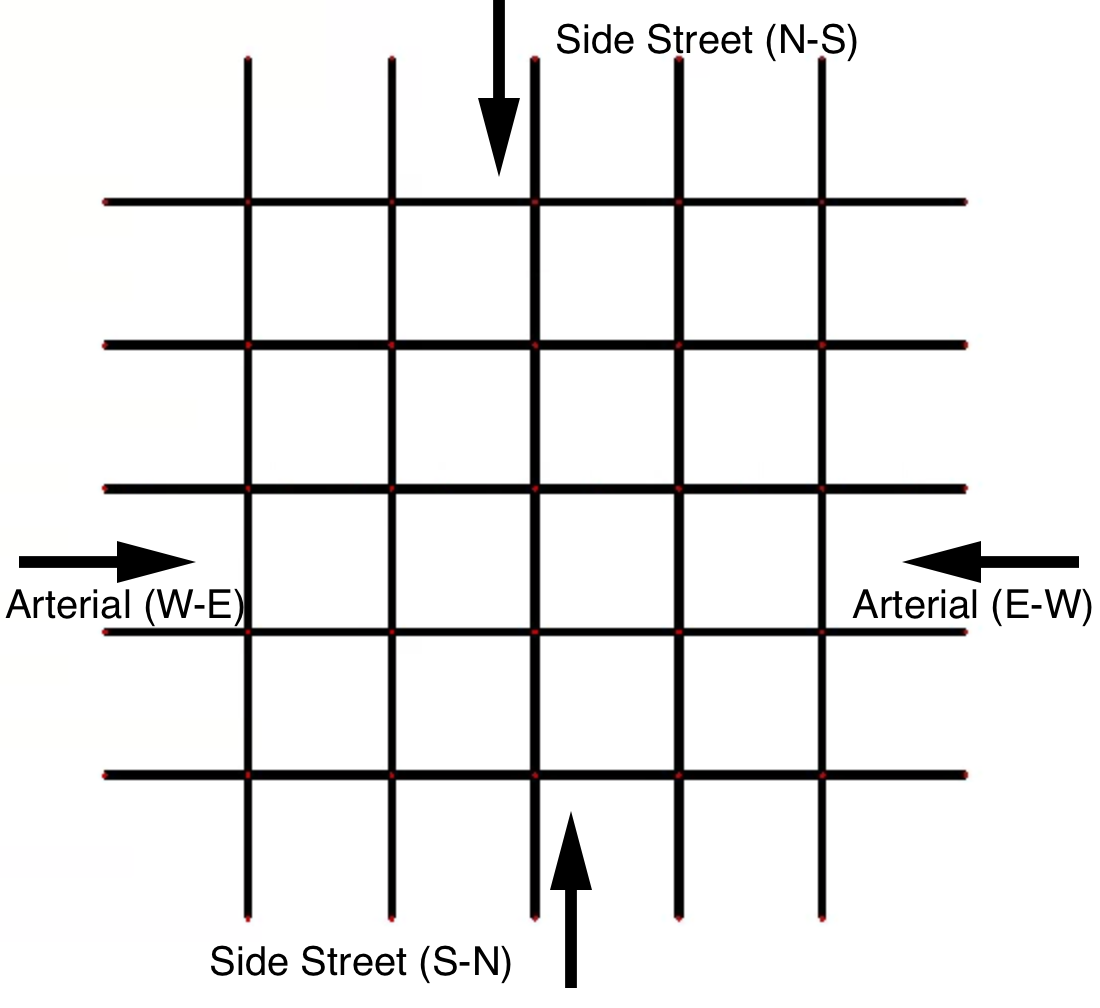}
	\end{minipage}}
\caption{Illustration of traffic demand patterns for pre-train and training}
\label{fig:scale_exp_flow}
\end{figure}

\begin{figure}[H]
\centering
\includegraphics[scale=0.7]{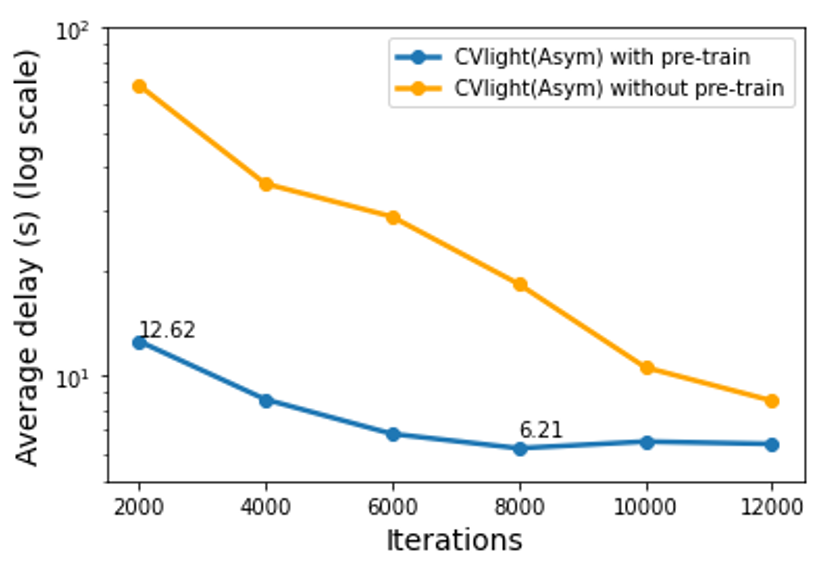}
\caption{\tcb{Comparison between CVLight (Asym) with and without pre-train in the convergence speed}}
\label{fig:pre-train-test}
\end{figure}

{\textbf{Comparison on convergence speed.} \tcb{We test the performance of the CVLight (Asym) agents under the 5-5 road network after every 2,000 iterations in training with or without the pre-train technique and the result is detailed in Figure \ref{fig:pre-train-test}. \tcb{ For Figure \ref{fig:pre-train-test}, the x-axis is the iteration index during training and the y-axis represents the average delay per vehicle per intersection in a logarithmic scale.} The blue line is the performance of CVLight (Asym) with the pre-train technique and the orange line is the performance of CVLight (Asym) without the pre-train technique. \tcb{ The numbers above the blue line are the average delays of CVLight (Asym) with pre-train at corresponding itertaions.} From the result, we can see CVLight (Asym) with pre-train can converge in a much faster way than that without using the pre-train technique does. We train our models on one c5a.xlarge AWS EC2 instance with 16 vCPUs and 32 GB memory. The processors are the 2nd generation 3.3GHz AMD EPYC 7002 series. 
% \tcb{By directly training all CVLight (Asym) agents for 12,000 iterations on the 5-5 road network, the training time is 190 minutes; by using the pre-train technique, the total training time is 150 minutes, which is composed of 55 minutes in pre-training and 95 minutes in the training on the 5-5 road network for 6,000 iterations.}
\tcb{By using the pre-train technique, the model begins to converge at the 6000th iteration, which is a $50\%$ improvement compared to the total 12000 iterations.}
}}

{ \textbf{Comparison on performance under 5-5.} \tcb{Under the 5-5 road network with 10\% and 30 \% penetration rates, the performance comparison between CVLight and baseline models is presented in Table \ref{tab:pre-train-results}. The first column of Table \ref{tab:pre-train-results} lists the algorithms and the second to third columns show the average delay as well as the standard deviation of each algorithm under the 10\% and 30\% penetration rates, respectively. From Table \ref{tab:pre-train-results}, CVLight (Asym) with pre-train can notably outperform all benchmark algorithms, which demonstrates the good scalability of CVLight.}}

% \begin{figure}[H]
% \centering
% \includegraphics[scale=0.3]{5-5_pretrain.png}
% \caption{CVLight (Asym) pre-train in 5-5 network} 

% \label{fig:pre-train-test}
% \end{figure}

\begin{table}[H]
\renewcommand{\thetable}{\arabic{table}}
\centering
\caption{\label{tab:pre-train-results} \tcb{Peformance comparison between CVLight (Asym) with pre-train and baselines in the 5-5 network}}
%\resizebox{\columnwidth}{!}{%
\begin{tabular}{cccccccccccc}
\toprule
\multicolumn{1}{c}{\multirow{2}{*}{\begin{tabular}[c]{@{}c@{}} Algorithms \end{tabular}}} & \multicolumn{3}{c}{CV penetration rate} \\ 
\cline{2-7}
\multicolumn{1}{c}{}  & 10\% & 30\% \\ 
\midrule
\makecell{CVLight (Asym) with pre-train}  & $\textbf{6.21}\pm0.62$ & $\textbf{5.43}\pm0.85$ \\
\midrule
\makecell{Presslight}  & $6.83\pm0.55$ & $5.54\pm0.68$  \\ 
\midrule
Maxpressure  & $7.85\pm0.95$ & $6.32\pm0.20$\\ 
\midrule
DQN  &  $8.70\pm0.64$ & $5.96\pm0.34$ \\ 
\midrule
Websters  &  $11.02\pm1.84$ & $7.73\pm1.43$ \\ 
\bottomrule
\end{tabular}
%}
\end{table}

\section{Case Study on Real-world Intersections}
\label{sec:case_study}

{In addition to synthetic intersections, we conduct experiments on a real-world road network of 4 intersections. In this section, we first introduce the new experiment set-up in Section \ref{subsec:case_study_settings} and then present experiment results in Section \ref{subsec:case_study_result}.}

%with various penetration rates

\subsection{Environment Set-up}
\label{subsec:case_study_settings}

\begin{figure}[H]
  \subfloat[Beaver Ave. and College Ave. from Atherton St. to Burrowes St., Pennsylvania, USA: a real-world 2-by-2 road network]{
	\begin{minipage}[c][1.06\width]{
	   0.5\textwidth}
	   \centering
	   \includegraphics[width=1.0\textwidth]{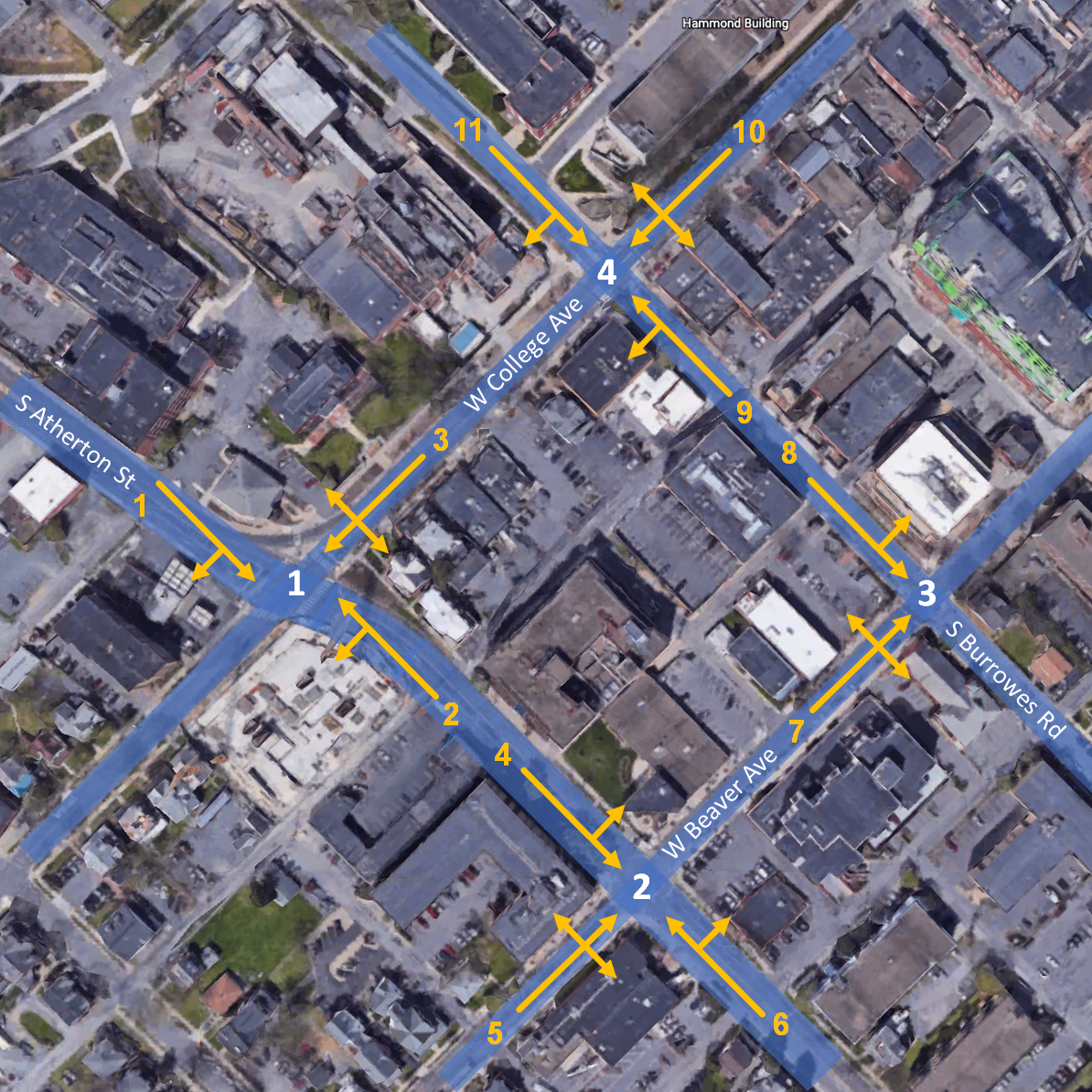}
	\end{minipage}}
 \hfill
  \subfloat[Traffic signal phases settings]{
	\begin{minipage}[c][1.06\width]{
	   0.5\textwidth}
	   \centering
	   \includegraphics[width=1.0\textwidth]{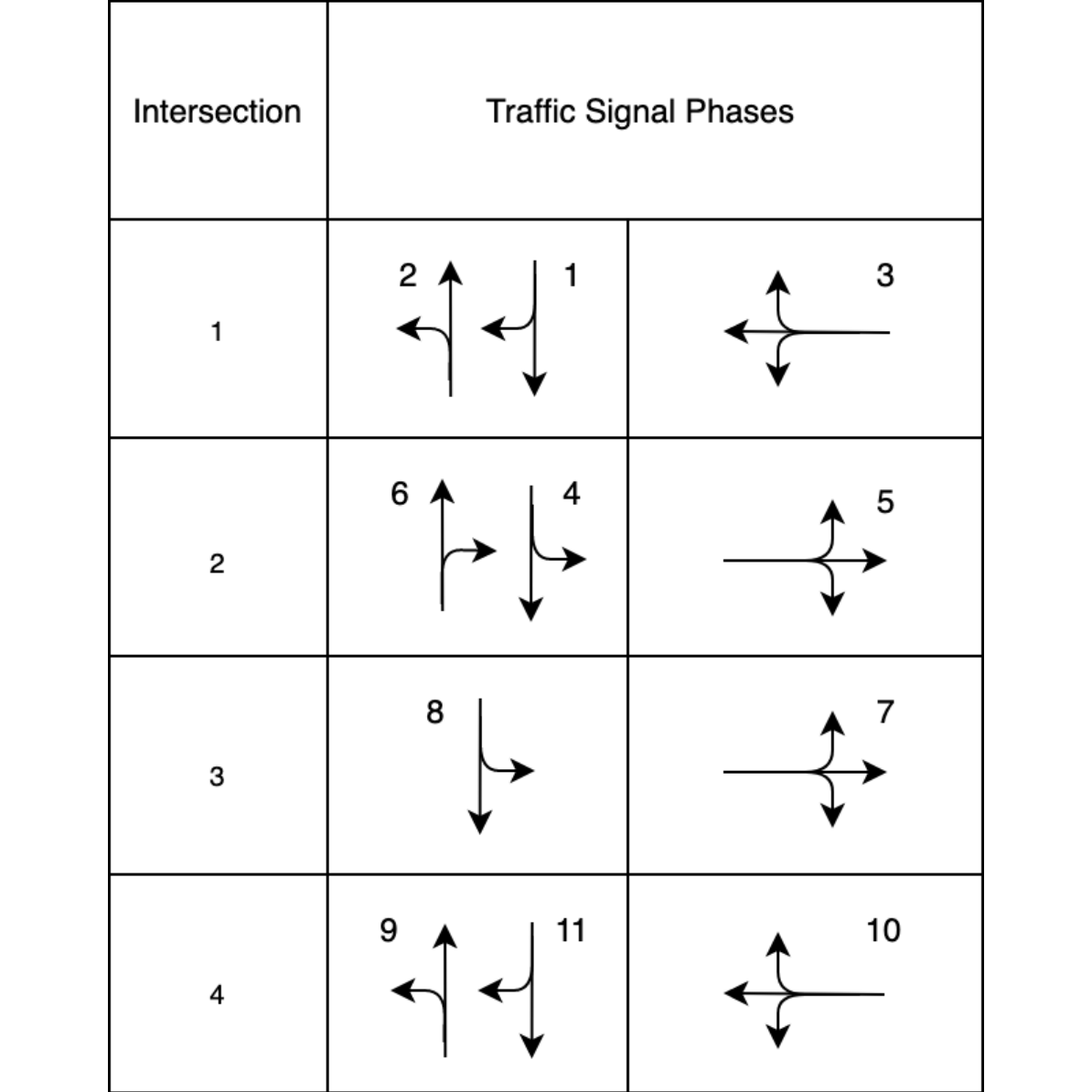}
	\end{minipage}}
\caption{Illustrations of intersection structure, traffic movements, and signal phase setting in the real-world road network}
\label{fig:illustrtaion_case_study}
\end{figure}

{Figure \ref{fig:illustrtaion_case_study} illustrates the intersection structure, traffic patterns,
%\tcr{(traffic flow is a non-countable noun and carries special meaning in the community. Here you refer to "traffic demand" or "traffic patterns.")}{(wz: traffic demands is used to represent the scale of arrival rate in this paper; to avoid causing confusion, I use ``traffic movement'' to represent the direction of traffic.)}
and traffic signal phase setting for the case study environment. Figure \ref{fig:illustrtaion_case_study}(a) shows the road network structure and traffic flows of a real-world 2-by-2 network located in the intersections of Beaver Avenue and College Avenue from Atherton Street to Burrowes Street in State College, Pennsylvania, USA. The Beaver Avenue and College Avenue are arterials with unidirectional traffic; the Atherton Street and the Burrowes Street are side-streets with bidirectional traffic. For a better comparison with PressLight~\citep{wei2019presslight}, we select this real-world road network because PressLight also gets tested under the corridor on the Beaver Avenue in the same area. Figure \ref{fig:illustrtaion_case_study}(b) details the traffic signal phase setting for each intersection in the road network. Each row represents the signal timing plan for one intersection and each cell inside one row indicates one green phase, consisting of traffic movements that are allowed for that phase.} 

{Vehicles' inter-arrival times follow a binomial distribution, with the same dynamic demand pattern shown in Table \ref{tab:traffic-demand} and Figure \ref{fig:dynamic_traffic_flow_design}. Both the left-turn proportion and right-turn proportion are set as 10\%. In this section, we focus on the performance comparison of CVLight (Asym) and other benchmark algorithms, considering the superior performance of CVLight (Asym) in Section \ref{sec:experiments}.}

\subsection{Result Analysis}
\label{subsec:case_study_result}

\begin{figure}[H]
\centering
\includegraphics[scale=0.9]{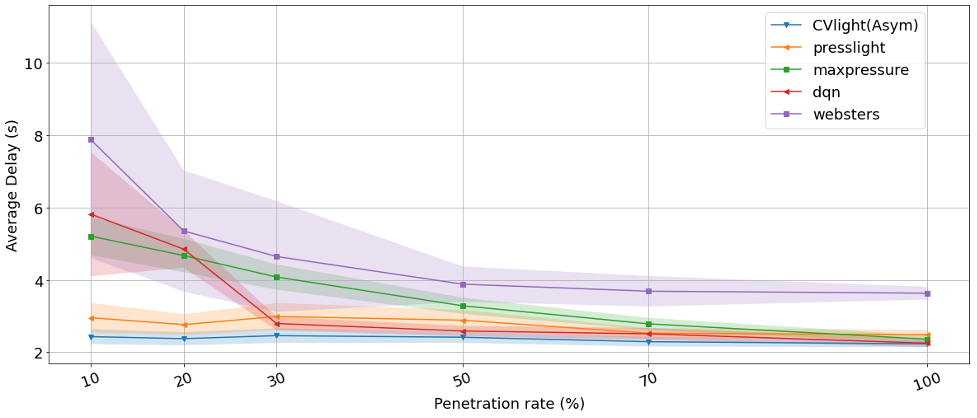}
\caption{Performance comparison under a real-world 2-by-2 road network} 

\label{fig:penn_state_comp}
\end{figure}

{Figure \ref{fig:penn_state_comp} illustrates the result of the performance comparison between CVLight (Asym) and baselines. The x-axis is the CV penetration rate and the y-axis represents the average delay per vehicle per intersection. The blue, orange, green, red, and purple lines represent the performance of CVLight (Asym), PressLight, Maxpressure, DQN, and Webster's, respectively. From Figure \ref{fig:penn_state_comp}, CVLight (Asym) can outperform all other benchmark algorithms across all penetration rates.}

%Our CVLight still performs poorly under extremely low penetration rates, like 10\%. 

%{Reviewer\#2 10. The results in Fig12 and Fig13 are confusing, as the performance of those benchmarks varies significantly, and the authors provide few explanations. Fig12, why max-pressure could outperform DQN and Presslight? Fig 13, why DQN is better than Presslight, or why PressLight deteriorates significantly in the corridor?}

%{Reviewer\#2 11. Overall, what makes the proposed CVLight outperform those of DQN and Presslight? Is that because of the dedicated state design or actor-critic structure with different inputs, or some factors else. The authors conduct extensive sensitivity analysis, but the main conclusion is mainly about the penetration rate of the CVs.}

\section{Conclusions and Future Research}
\label{sec:discussions&conclusion}

{ This paper develops the CVLight model that learns with information from CVs and non-CVs and executes with data collected from CVs only. The model is characterized by the state and reward design that facilitates coordination among agents and utilizes travel delay and phase duration. We propose the Asym-A2C that takes advantage of the structure of actor-critic algorithm so as to include non-CV information into the training process. Results of extensive experiments under various traffic demands, penetration rates, as well as road networks, demonstrate the superiority of CVLight over other state-of-the-art benchmark algorithms and its good generalizability in penetration rates and traffic demands. By visualizing the learned policies of CVLight agents and the relative importance of input layer neurons in the critic neural network, we show how the Asym-A2C algorithm and our state design contribute to the good performance of CVLight (Asym), especially under low penetration rate scenarios. Moreover, the scalability of CVLight is further improved through the pre-train technique, which significantly decreases the training time of CVLight under a 5-by-5 road network.}

{Future work is threefold}:
(1) {We plan to explore more state-of-the-art neural network interpretability methods to better interpret the learned policies of agents.} (2) {We will further improve the scalability of CVLight by introducing other deep learning methods, such as parameter sharing.}
(3) {Scenarios like extremely low CV penetration can still be challenging for TSC. Following the suit of some existing studies, we will combine state estimation methods, such as queue estimation, with RL-TSC so as to further improve our model performance. }

\section*{Acknowledgements}
This work is partially sponsored by the National Science Foundation (NSF) under the CAREER award CMMI-1943998   
and NSF CPS-2038984.
%the Region 2 University Transportation Research Center (UTRC) (subcontract RUTGER PO 966112/PID824227). 
We also thank Mr. Ujwal Dinesha and Yaxing Cai for assisting in coding in the first phase of this project. 

\bibliographystyle{elsarticle-harv}
\bibliography{main}

\end{document}